%% file: main.tex
\documentclass[acmsmall,screen,nonacm]{acmart}

\newcommand{\ed}[1]   {\textcolor{black}{#1}}
\newcommand{\add}[1]   {\textcolor{black}{#1}}

\newcommand{\edreview}[1]   {\textcolor{black}{#1}}
\newcommand{\addreview}[1]   {\textcolor{black}{#1}}

\AtBeginDocument{%
  \providecommand\BibTeX{{%
    \normalfont B\kern-0.5em{\scshape i\kern-0.25em b}\kern-0.8em\TeX}}}

\setcopyright{acmlicensed}
\acmJournal{CSUR}
\acmYear{2022} \acmVolume{1} \acmNumber{1} \acmArticle{1} \acmMonth{1} \acmPrice{15.00}\acmDOI{10.1145/3527450}

\begin{document}

\title[Software-Based Dialogue Systems: Survey, Taxonomy and Challenges]{Software-Based Dialogue Systems: Survey, Taxonomy and Challenges}

\author{Quim Motger}
\email{jmotger@essi.upc.edu}
\author{Xavier Franch}
\email{franch@essi.upc.edu}
\affiliation{%
  \institution{Department of Service and Information System Engineering (ESSI), Universitat Politècnica de Catalunya (UPC)}
  \country{Spain}
}

\author{Jordi Marco}
\affiliation{%
  \institution{Department of Computer Science (CS), Universitat Politècnica de Catalunya (UPC)}
  \country{Spain}
  }
\email{jmarco@cs.upc.edu}


\label{abstract}
\input{content/0_abstract}



\maketitle

\label{introduction}
\input{content/1_introduction}

\label{researchMethod}
\input{content/2_researchMethod}

\input{content/3_1_published_research}
\input{content/3_2_hci_features}
\input{content/3_3_technical_methods}
\input{content/3_4_training_testing_eval}
\input{content/3_5_future_challenges}



\label{limitations}
\input{content/5_limitations}

\label{conclusions}
\input{content/6_conclusions}

\label{acknowledgments}
\input{content/A_acknowledgments}

\bibliographystyle{ACM-Reference-Format}
\bibliography{base}


\end{document}

%% file: content/0_abstract.tex

\begin{abstract}

The use of natural language interfaces in the field of human-computer interaction is undergoing intense study through dedicated scientific and industrial research.
The latest contributions in the field, including deep learning approaches like recurrent neural networks, the potential of context-aware strategies and user-centred design approaches, have brought back the attention of the community to software-based dialogue systems, generally known as conversational agents or chatbots. Nonetheless, and given the novelty of the field, a generic, context-independent overview on the current state of research of conversational agents covering all research perspectives involved is missing.
Motivated by this context, this paper reports a survey of the current state of research of conversational agents through a systematic literature review of secondary studies.
The conducted research is designed to develop an exhaustive perspective through a clear presentation of the aggregated knowledge published by recent literature within a variety of domains, research focuses and contexts.
As a result, this research proposes a holistic taxonomy of the different dimensions involved in the conversational agents' field, which is expected to help researchers and to lay the groundwork for future research in the field of natural language interfaces.

\end{abstract}

%% file: content/1_introduction.tex
\section{Introduction}

\textit{Conversational agents}, \textit{chatbots}, \textit{dialogue systems} and \textit{virtual assistants} are some of the terms used by scientific literature \cite{Syvanen2020-nb} to describe software-based systems which are capable of processing natural language data to simulate a smart conversational process with humans \cite{Dale2016}. These conversational mechanisms are built and driven by a wide variety of techniques of different complexity, from traditional, pre-coded algorithms to emerging adaptive machine learning algorithms \cite{Neff2016_np}. Usually deployed as service-oriented systems, they are designed to assist users to achieve a specific goal based on their personal needs \cite{Folstad2017-bp}. To this end, they autonomously generate natural language messages to interact and communicate with users by emulating a real human being \cite{Riikkinen2018_ss, GRIOL201927}.

Although the interest in both industry and research has dramatically increased in recent years \cite{Maedche2019}, the study of natural language communication between human beings and machines is indeed not a novel concept. ELIZA \cite{Weizenbaum1983}, which has been historically considered as the first chatbot, was designed and developed by the Massachusetts Institute of Technology (MIT) more than half a century ago, between 1964 and 1966. Alongside successive chatbots like PARRY \cite{parry1972}, these innovative systems laid the groundwork for specialized research in the field of human-computer interaction (HCI), focusing on the social and communicative perspectives and their impact on the design and development of these systems. 
Recent advances in the field of artificial intelligence have brought back attention to the potential of conversational agents, especially with the emergence of machine and deep learning techniques \cite{Bavaresco2020-md}. Furthermore, specialized research fields such as natural language understanding (NLU), natural language generation (NLG) and dialogue stage tracking (DST) have become disruptive areas by introducing innovative, efficient and accurate solutions to machine cognitive problems \cite{WANG2016303}. 

Consequently, several literature reviews and surveys have been conducted in recent years \cite{Hussain2019-lj, Knote2018-dx}, with an especial emphasis on the implications of the latest innovations from a technical perspective \cite{Dsouza2019-px, Akma2018-hz}. Nevertheless, these secondary studies are typically limited in terms of context or domain of application \cite{Gentner2020-ns}, focus of research field \cite{Palasundram2020-mk} or even research method \cite{Ahmad2018-at}. 
To the best of our knowledge, no literature review in the field of conversational agents addresses all the research dimensions involved in this field.

All things considered, in this paper, we conduct a state-of-the-art review in the field of conversational agents, and their impact in the software engineering field, applying a rigorous research method based on Kitchenham et al. guidelines on systematic reviews \cite{Kitchenham2010-wk}. The contributions of this work are: to provide an up-to-date, holistic review of the conversational agents' research field; to present a taxonomy of the main concepts uncovered in the study, which will allow researchers in the field to classify their work and compare to other's proposals; to identify the main challenges and research directions for future work. 

The remainder of this article is organized as follows. Section \ref{sec:research-method} describes the research method. 
Sections \ref{sec:sq1}, \ref{sec:sq2}, \ref{sec:sq3}, \ref{sec:sq4} and \ref{sec:sq5} present the results of the feature extraction process and the conclusions of each of the research questions. 
Section \ref{sec:limitations} evaluates the limitations of the study. Finally, Section \ref{sec:conclusions} summarizes the main conclusions.

%% file: content/2_researchMethod.tex
\section{Research method}
\label{sec:research-method}

We conducted our survey as a tertiary study over the scientific literature in the domain. A tertiary study is a systematic literature review used to find and synthesize the knowledge in a wide scope of research that is scattered in existing reviews. We followed Kitchenham et al.'s approach to perform tertiary studies \cite{Kitchenham2010-wk} with a few modifications, remarkably:

\begin{itemize}
    \item In addition to automated search selection, we conducted a reference snowballing process to avoid missing relevant papers in the field of conversational agents (see Section \ref{subsec:snowballing}).
    \item We extended the quality assessment criteria list according to the latest practices in the software engineering field (see Section \ref{subsec:quality-analysis}).
\end{itemize}
 
The detailed research method and the generated artefacts are collected and published in a replication package\footnote{Available at: \href{https://zenodo.org/record/5718617}{https://zenodo.org/record/5718617}}, which allows the auditability and the replicability of the research process.
 
\subsection{Research questions}
\label{subsec:research-questions}

This paper addresses the following general research question (GQ):

\begin{itemize}
    \item \textit{\textbf{GQ.} What is the current state of research in the field of conversational agents?}
\end{itemize}

This general question is refined into a sub-set of specific scientific questions (SQs). 
Each SQ is supported through the feature extraction process by categorizing the sub-set of studies addressing the topic covered by each SQ (see Section \ref{subsec:data-extraction}). These SQs are defined as follows:

\begin{itemize}
    \item \textit{\textbf{SQ1.} What research has been published in the field of conversational agents?}
    \item \textit{\textbf{SQ2.} Which HCI features have a relevant impact on user experience with conversational agents?}
    \item \textit{\textbf{SQ3.} Which technical methods and technologies are used for the design and implementation of conversational agents?}
    \item \textit{\textbf{SQ4.} Which methodological approaches and resources are used for training, testing and evaluating conversational agents?}
    \item \textit{\textbf{SQ5.} Which research challenges are reported by conversational agents' literature?}
\end{itemize}

\subsection{Search string}

Based on our general research question, two key major concepts are identified: \textit{state of the art} and \textit{conversational agents}. Regarding the former, and since we have adopted tertiary studies as our methodological approach, we decided to include literature reviews performed according to a rigorous process. To guarantee that the search exhaustively covers all relevant literature in the field, we use a set of synonyms as proposed by Kitchenham et al. \cite{Kitchenham2010-wk} which encompass the majority of structured literature reviews and secondary studies (e.g., literature reviews, mapping studies, literature surveys...). Regarding the latter, we carried on an exploratory review of the current state of research through recent systematic literature reviews in the field to include an extensive selection of alternative terms used by the literature that refer to \textit{conversational agents} (e.g., chatbots, dialogue systems, conversational entities...). \add{The search string is depicted in detail in Table \ref{tab:search-string}}.

\begin{table}[h]
  \caption{\add{Search string}}
  \label{tab:search-string}
  \begin{tabular}{p{0.15\linewidth}p{0.80\linewidth}}
    \toprule
    \textbf{\add{Major term}}&\textbf{\add{Search string}}\\
    \midrule
    \add{Literature review}&\add{("review of studies" OR "structured review" OR "systematic review" OR "literature review" OR “mapping review” OR “mapping study” OR "literature analysis" OR "in-depth survey" OR "literature survey" OR "meta-analysis" OR "past studies" OR "subject matter expert" OR "analysis of research" OR "empirical body of knowledge" OR "overview of existing research" OR "body of published research")}\\
    &\add{AND}\\
    \add{Conversational agent}&\add{("conversational agent*" OR "dialogue system*" OR "intelligent virtual assistant*" OR "intelligent virtual agent*" OR "relational agent*" OR "intelligent personal assistant*" OR "intelligent personal attendant*" OR "automated personal attendant*" OR "personal assistant*" OR "personal attendant*" OR "chat-bot*" OR "chatbot*" OR "chat bot*" OR "chatterbot*" OR "artificial conversational entit*")}\\
    \bottomrule
\end{tabular}
\end{table}

\subsection{Data sources}
\label{subsec:ds}

We selected the following digital libraries as well-known, broadly used repositories in literature reviews from the computer science and software engineering fields: Web of Science
, Scopus
, ACM Digital Library
, IEEE Xplore
, Science Direct 
and Springer Link.

\subsection{Study selection strategy}
\label{subsec:study-selection-strategy}

\begin{figure}[h]
  \centering
  \includegraphics[width=\linewidth]{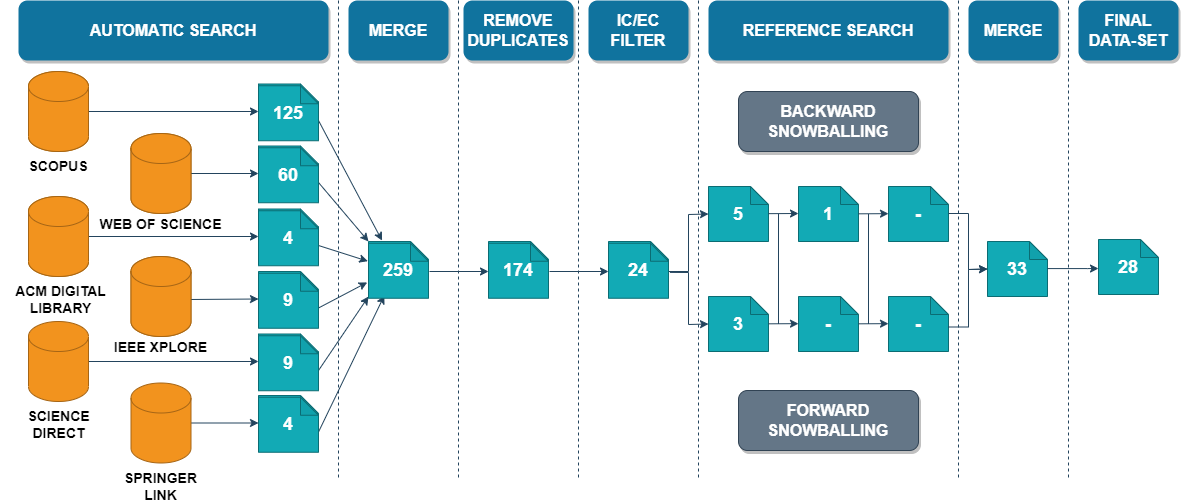}
  \caption{Summary of the study selection process.}
  \label{fig:study-selection}
\end{figure}

In this section, we depict the study selection strategy, whose results are summarized in Figure \ref{fig:study-selection}. 

\subsubsection{Study search}
\label{subsubsec:study-search}

The study search is designed as the following set of semi-automatic tasks:

\begin{enumerate}
    \item Customize the search string to the 
    syntactic requirements of each data source.
    \item Query\footnote{The search was conducted \edreview{for all studies published before} the 31st January, 2021} the adapted string in each data source engine using the filter by title, abstract and authors' keywords. No restriction regarding the date of publication was applied.
    \item Export the bibliographic references of the search results into a shared spreadsheet document.
    \item Integrate all references using a common format based on all shared fields.
    \item Filter out and remove all duplicated references.
\end{enumerate}

This process resulted in an initial data-set of \textbf{\ed{174} studies}.

\subsubsection{Inclusion and exclusion criteria}
\label{subsec:ic-ec}

After the extraction of the initial data-set of studies from the study selection process, we applied a set of inclusion and exclusion criteria to evaluate each of the selected studies. For the inclusion criteria (IC), we include those studies matching all the following:

\begin{itemize}
    \item \textbf{IC1.} The paper is a literature review as defined by Kitchenham et al. \cite{Kitchenham2010-wk}.
    \item \textbf{IC2.} The focus of the research is on conversational agents, chatbots or any of the synonyms included in the exploratory analysis.
\end{itemize}

Regarding the exclusion criteria (EC), we exclude all studies matching any of the following:

\begin{itemize}
    \item \textbf{EC1.} Research results, contributions and conclusions cannot be applied to the computer science, software engineering or artificial intelligence research fields. 
    \item \textbf{EC2.} Research focuses exclusively on embodied conversational agents, virtual reality or 3-D modelling features of conversational agents.
    \item \textbf{EC3.} The paper is a previous study of a more recent publication which has already been included in the initial data-set.
    \item \textbf{EC4.} The paper is a work in progress.
    \item \textbf{EC5.} Full-text is not available and it has not been delivered after request to their authors.
    \item \textbf{EC6.} Full-text is not published in English.
\end{itemize}

From the initial data-set of \ed{174} studies, only \ed{81} matched both ICs. Concerning the most common ECs, \ed{54} studies were not focused on analysing and reporting research conclusions and results directly related to the computer science domain (EC1). Some examples include analysis on healthcare rhetorical structure theory \cite{Hou2020-xf}, digital health coaching programs \cite{Stara2020-ky} or behavioural change in patients using medical chatbots \cite{Gentner2020-ns}. On a secondary basis, up to 11 studies matching EC2 were focused on embodied conversational agents \cite{Stal2020-vm, Lopez2007-kf}  or physical modelling features \cite{Blut2021-ip}. Finally, we still excluded \ed{19} additional studies applying the rest of ECs.


As a result, we identified a sub-set of \textbf{24 studies} fulfilling all ICs and none of the depicted ECs. 

\subsubsection{Reference snowballing}
\label{subsec:snowballing}

To extend the scope of research and to minimize the risk of missing relevant studies, we integrated an additional reference search strategy into the study selection strategy. The snowballing reference search process is designed as depicted by Wohlin \cite{Wohlin2014-ks} based on iterative backward and forward reference searches. In each iteration, we conducted a backward and forward snowballing process of all secondary studies included in our data-set. Each reference fitting the goal of this research and passing the IC/EC evaluation (i.e., matching all IC and none of the EC defined in Section \ref{subsec:ic-ec}) was added to the original data-set, and a new iteration was applied to the enlarged data-set. We conducted three complete iterations until reaching saturation (see Figure \ref{fig:study-selection}). This process resulted in \ed{10} new studies, which led to a total amount of \textbf{\ed{33} studies}.

\subsubsection{Quality assessment}
\label{subsec:quality-analysis}

The study selection step concluded with a quality assessment to evaluate the quality of the research strategies from the selected secondary studies. Using relevant tertiary studies from the software engineering field as reference \cite{Cadavid2020-xk, Kudo2019-gz, Xu2020-qw}, the quality assessment plan was designed based on the quality assessment criteria (QAs) included in the DARE-5 extended proposal from the York University Centre for Reviews and Dissemination \cite{DARE5} as refined by Kitchenham et al. in systematic reviews \cite{Kitchenham2009-aq} and tertiary studies \cite{Kitchenham2010-wk}. We define these QAs as follows:

\begin{itemize}
    \item \textbf{QA1.} Does the search protocol presumably cover all relevant data sources?
    \item \textbf{QA2.} Are inclusion and exclusion criteria adequately described?
    \item \textbf{QA3.} Does the search protocol include a quality assessment evaluation?
    \item \textbf{QA4.} Are primary studies adequately identified and described?
    \item \textbf{QA5.} Is there a synthesis process to summarize and support research results and conclusions?
\end{itemize}


We used this QAs to compute a global quality score for each paper to remove all secondary studies with a quality score of 0 (details on QAs evaluation are available in the replication package).
After applying this QA filter, we filtered out 5 studies, which gave us a final data-set of \textbf{\ed{28} studies}. 

\subsection{Data extraction}
\label{subsec:data-extraction}

The data extraction process was designed in alignment with the research questions described in Section \ref{subsec:research-questions}. We identified and defined a set of research data features (F), which are intended to exhaustively cover the scope and topics covered by the SQs included in our research while facilitating the dissertation and discussion.

\begin{itemize}
    \item \textbf{SQ1 $\rightarrow$ F1. Terminology.} Identification and analysis of the terminology used by literature to describe and refer to conversational agents, including synonyms, variants and descriptors.
    \item \textbf{SQ1 $\rightarrow$ F2. Domains.} Identification and categorization of business domains and areas of applications covered by research in the conversational agents' field.
    \item \textbf{SQ1 $\rightarrow$ F3. Goals.} Identification and categorization of the primary goals of integrating conversational agents into software systems.
    \item \textbf{SQ2 $\rightarrow$ F4. HCI features.} Identification and categorization of HCI features and specifications of conversational agents, including verbal communication, non-verbal communication and appearance features. Evaluation of the impact and relevance of such features in the literature.
    \item \textbf{SQ3 $\rightarrow$ F5. Design dimensions.} Enumeration and depiction of high-level software architecture characteristics for integrating conversational agent into software systems.
    \item \textbf{SQ3 $\rightarrow$ F6. Technical implementation specifications.} Enumeration and depiction of technical development features, specifications and technologies for conversational agents.
    \item \textbf{SQ3 $\rightarrow$ F7. Context integration techniques.} Enumeration and depiction of strategies for the integration of contextual data, personalization and self-evolution of conversational agents.
    \item \textbf{SQ4 $\rightarrow$ F8. Data-sets and data items.}  Enumeration and depiction of natural language data-sets used for training, testing and evaluating conversational agents in primary studies.
    \item \textbf{SQ4 $\rightarrow$ F9. Quality and evaluation methods.} Identification and categorization of the quality features, evaluation methods and metrics used for quality analysis of primary studies.
    \item \textbf{SQ5 $\rightarrow$ F10. Research challenges.} Enumeration and depiction of future research challenges, trends and potential emerging strategies for the successful integration and evolution of the conversational agents' research field.
\end{itemize}

\input{content/4_researchMethodFindings}

%% file: content/4_researchMethodFindings.tex
\subsection{\ed{Research method findings}}
\label{sec:sq6}


\ed{To conclude the presentation of the research method, we summarized the main findings of the research strategy depicted in this survey. The purpose of this report is two-fold. First, to reinforce and validate the research method depicted in this section. And second, given the highly evolving nature of the covered research area, we envisage that future research will require from up-to-date surveys in the field, for which we expect that the lessons learned from this research can serve as an input for future researchers.}

\begin{figure}[h]
  \centering
  \minipage{0.60\textwidth}
      \includegraphics[width=\linewidth]{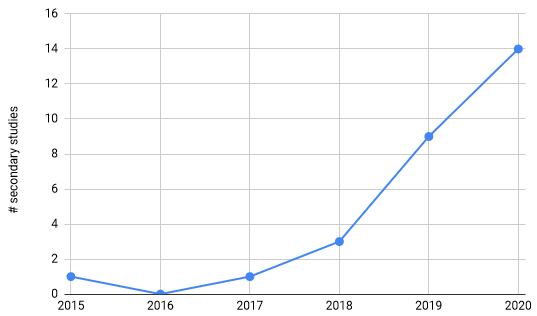}
      \caption{\ed{Distribution of secondary studies per year of publication}}
      \label{fig:year}
  \endminipage\hfill
  \minipage{0.36\textwidth}
      \includegraphics[width=\linewidth]{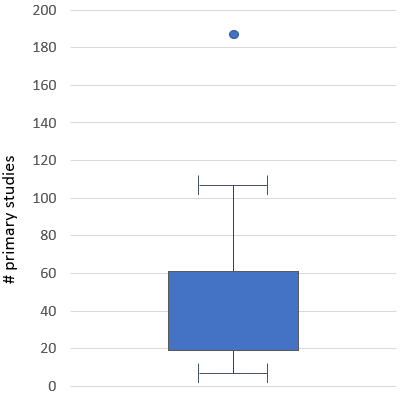}
      \caption{\ed{Distribution of primary studies}}
      \label{fig:ps}
  \endminipage\hfill
\end{figure}

\noindentparagraph{\textbf{\ed{Bibliometrics.}}}
\ed{Based on the final data-set of selected publications, we analyse two bibliometric metadata features: distribution of secondary studies per year of publication (Figure \ref{fig:year}), and distribution of the number of primary studies covered by each secondary study (Figure \ref{fig:ps}). Year of publication distribution reports an incipient trend in the conversational agents' field, which is reflected in an increasing growth of the publication of secondary studies in recent years. Only one of the studies \cite{Abdul-Kader2015-jy} was published before 2016, while the remaining 27 are comprised between 2017 and 2020. On the other hand, Figure \ref{fig:ps} reports the distribution of the number of primary studies covered by the included literature reviews. This increasing tendency of published literature reviews and the significant amount of primary studies included by these studies demonstrate that conversational agents are a subject undergoing intense research. Additionally, it also reinforces our selection of tertiary studies as the research method for conducting this survey}.  

\noindentparagraph{\textbf{\ed{Search strings.}}} 
\ed{We used the search string depicted in the surveyed secondary studies when explicitly depicted (21/28) 
as proxy for this analysis.
As a result, we find a clear evidence on the prominent use of \textit{conversational agent} and \textit{chatbot} as synonyms in research strategies, which is reinforced by the findings on the terminology (F1). Both terms are predominantly included in the majority of secondary studies' search strings (13/21, 62\%), and almost all of them include at least one of these terms (19/21, 90\%). 
Therefore, while typically researchers focus on using a single term in their dissertation for consistency, the majority of research strategies covered by this study include both \textit{chatbot} and \textit{conversational agent} in their research method. Alternative terms like \textit{chatterbot}, \textit{conversational interface} or \textit{dialogue system} are also frequently used, but they have a significant minor presence. Below this frequency rate, more than 40 synonyms are used in two or fewer studies, including chatbot variants (e.g., \textit{chat-bot}, \textit{chatterbox}), agent variants (e.g., \textit{embodied agent}, \textit{personal attendant}) or conversation variants (e.g., \textit{conversational system}, \textit{conversational assistant}). This exhaustive analysis of the search strings is useful for researchers performing search-based literature reviews in the domain, as it allows to define the search string more accurately.}

\begin{figure}[h]
  \centering
  \minipage{0.425\textwidth}
      \includegraphics[width=\linewidth]{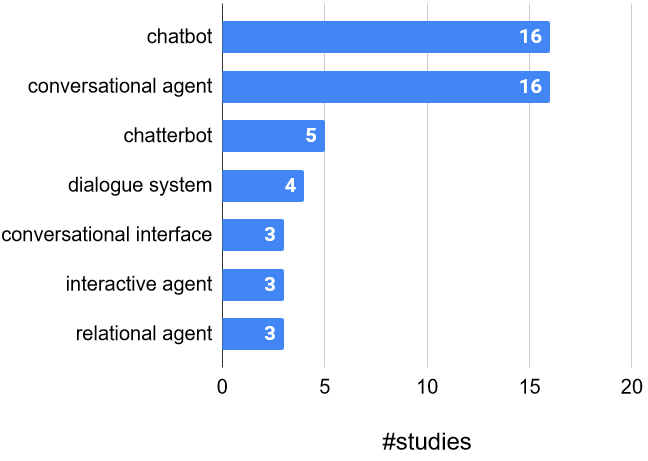}
      \caption{\ed{Terms used by secondary studies}}
      \label{fig:string-1}
  \endminipage\hfill
  \minipage{0.45\textwidth}
      \includegraphics[width=\linewidth]{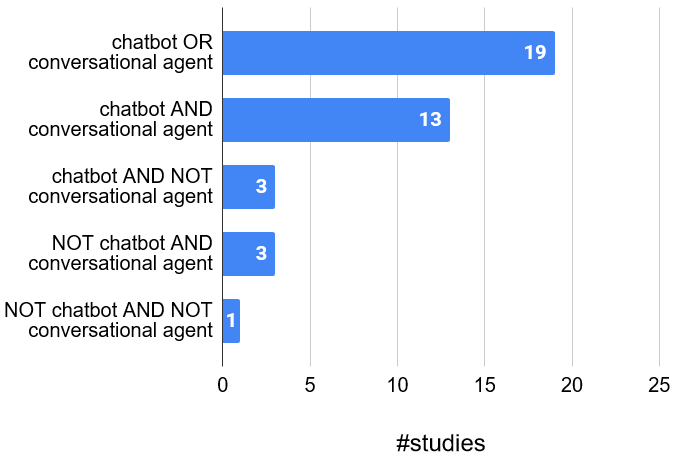}
      \caption{\ed{Combined use of main terms}}
      \label{fig:string-2}
  \endminipage\hfill
\end{figure}

\begin{figure}[h]
  \centering
  \minipage{0.45\textwidth}
      \includegraphics[width=\linewidth]{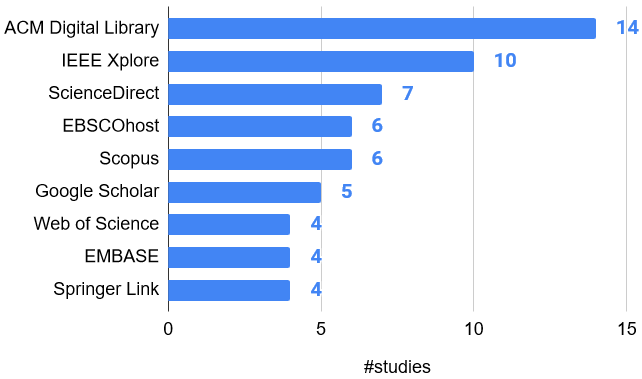}
      \caption{\ed{DLs used by secondary studies}}
      \label{fig:db}
  \endminipage\hfill
  \minipage{0.45\textwidth}
      \includegraphics[width=\linewidth]{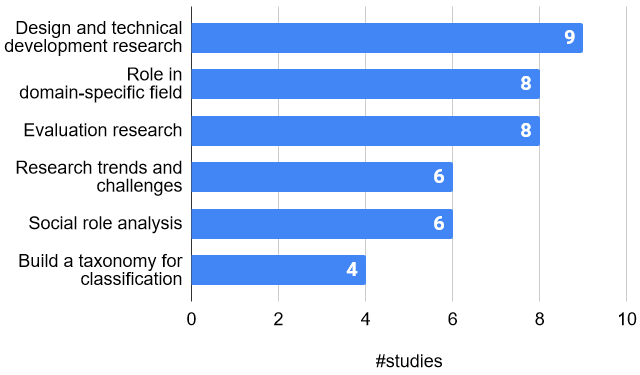}
      \caption{\ed{Secondary studies research purposes}}
      \label{fig:goals}
  \endminipage\hfill
\end{figure}

\noindentparagraph{\textbf{\ed{Digital libraries.}}}
\ed{Regarding the digital libraries (Figure \ref{fig:db}) used for automated search (explicitly covered by 19/28 secondary studies), the most common database is ACM Digital Library, followed by IEEE Xplore, Science Direct, EBSCOhost, Scopus, Google Scholar, Web of Science, EMBASE and Springer Link. With a frequency rate below 4 studies we mostly observe domain-specific databases (e.g., PubMed) or publisher databases whose results are generally covered by other databases like Scopus. The use of less popular digital libraries like EBSCOhost and EMBASE can be justified by the relevance of related literature in the field of healthcare and education, as well as dedicated research on the social implications of the use of chatbots.} 


\noindentparagraph{\textbf{\ed{Research purpose.}}}

\ed{We inductively categorize the secondary studies under six classes, based on the focus of the research. The most common research goal is (1) the analysis of design and technical implementation of software systems based on the integration of conversational agents (9/28, 32\%). 
Discussion on (2) the role of conversational agents and their design features in a domain-specific field (8/28, 33\%) and (3) research on quality evaluation characteristics, methods and metrics (8/28, 33\%)
are the second most common goals. 
The remaining research goal categories are less frequent and include: (4) research trends and challenges, (5) social and communicative features analysis and (6) a taxonomy proposal for classifying conversational agents based on high-level design dimensions. The proposed classification for the research goals of the surveyed studies is exhaustively covered by the scientific research questions and the features of this research. Specifically: design and technical development research (SQ3), role in domain-specific field (SQ1), evaluation research (SQ4), research trends and challenges (SQ5), social role analysis (SQ2) and a taxonomy for classification (SQ1, SQ2, SQ3, SQ4).}

\noindentparagraph{\textbf{\addreview{Publication venues.}}}

\addreview{The primary studies referenced by the literature reviews in our surveyed data-set are significantly scattered among a wide variety of interdisciplinary conferences and journals from different scientific research fields where conversational agents are becoming pervasive systems. Beyond purely computing-oriented venues, these include research fields like healthcare (e.g., \textit{Journal of Medical Internet Research}), engineering (e.g., \textit{Expert Systems with Applications}), psychology (e.g., \textit{Computers in Human Behavior}) and artificial intelligence (e.g., \textit{Conference on Artificial Intelligence (AAAI)}). Concerning top tier venues in the field of dialogue systems, we highlight the presence of well-reputed venues like the \textit{Association for Computational Linguistics} (ACL) \cite{qiu-etal-2018-transfer, galitsky-ilvovsky-2017-chatbot, hancock-etal-2019-learning, Lopez2007-kf, Cui2017-pn, Qiu2017-lw} and the \textit{Special Interest Group on Discourse and Dialogue} (SIGdial) \cite{Kobayashi2015-le, Morbini2012, brixey-etal-2017-shihbot}. However, it is significant to highlight that, given that the research method has been depicted in the form of a survey of secondary studies, it is possible that the surveyed studies do not include some relevant, top tier venues in the field which do not publish this kind of studies, but mainly focus on publishing primary studies. 
}

%% file: content/3_1_published_research.tex
\section{Published research (SQ1)}
\label{sec:sq1}

\subsection{Terminology (F1)}
\label{subsec:f1}

Being aware of the use of different synonyms in the literature to denote the concept of conversational agents, we investigated in more depth which are these terms and how are they used. The surveyed publications show a lack of dedicated discussion concerning terminology, and there is no taxonomy of reference which is generally used by researchers. While \textit{conversational agent} and \textit{chatbot} are the most common terms, it is not clear that there is an evident difference between these concepts. Radziwill and Benton \cite{Radziwill2017-ws} present a classification of software-based dialogue systems, where \textit{conversational agents} are a sub-class of \textit{dialogue systems}, and \textit{chatbots} and \textit{embodied conversational agents} are both sub-classes of \textit{conversational agents}. However, this three-level hierarchy is not representative of the findings of this survey. In fact, Syvänen and Valentini \cite{Syvanen2020-nb} demonstrate that terms like \textit{chatbot}, \textit{virtual assistant}, \textit{agent}, \textit{conversational bot} or even \textit{robot} are used without an explicit definition in the majority of studies. And when a definition is provided, they are generally used indistinctly among literature with varied definitions, whose differences are not significant. Moreover, a detailed analysis of the terminology used by the studies included in this research demonstrates that only a few studies use the term \textit{chatbot} as a different term (i.e., a sub-class) of \textit{conversational agent} \cite{VanPixteren2020-zm, Radziwill2017-ws, Masche2018-sw}. While in some studies the term \textit{conversational agent} and its variants are used as descriptors to define \textit{chatbots} \cite{Masche2018-sw, Safi2020-cy}, the wide majority either use them indistinctly \cite{Bavaresco2020-md, Hussain2019-lj} or they present them as synonyms using a shared definition \cite{Rheu2021-fq, Perez2020-xt, Nuruzzaman2018-iv,Dsouza2019-px}. 

If we focus on these definitions and the descriptors used, they are generally described as computer programmes \cite{Syvanen2020-nb}, tools \cite{Perez2020-xt} or software systems \cite{Nuruzzaman2018-iv, Feine2019-vc} which are capable of simulating a human conversation \cite{VanPixteren2020-zm, Rheu2021-fq, Radziwill2017-ws} using natural language \cite{Knote2018-dx, Hussain2019-lj} or even artificial intelligence techniques \cite{Perez2020-xt, Milne-Ives2020-zl}. Additionally, they are also occasionally described as voice or text interfaces \cite{Perez2020-xt, Gabarron2020-gj} providing access to a set of services or functionalities \cite{Radziwill2017-ws, Janssen2020-fp}. All in all, for the sake of consistency and clarity, we propose to build a concise but exhaustive shared definition of \textit{conversational agents} and \textit{chatbots}, which will be used as synonyms in this research. \\

\noindent
\fbox{
  \parbox{0.97\linewidth}{
    \textit{\textbf{Conversational agents}} or \textit{\textbf{chatbots}} are software-based dialogue systems designed to simulate a human conversational process by processing and generating natural language data through a text or voice interface to assist users in achieving a specific goal \add{or satisfying a specific need}.
  }
}\\

\subsection{Domains (F2)}
\label{subsec:f2}

\begin{figure}[h]
  \centering
  \includegraphics[width=0.90\linewidth]{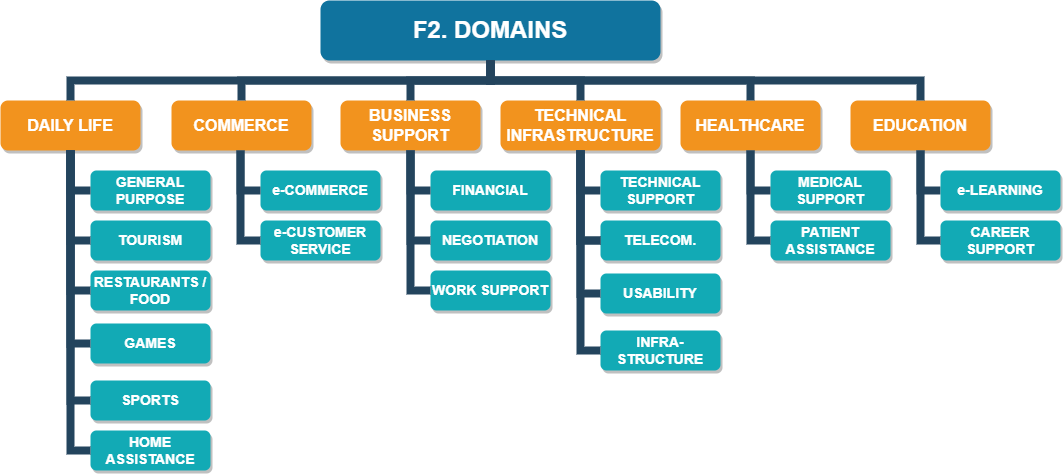}
  \caption{Aggregated results of conversational agents' domains in related literature}
  \label{fig:f2-sum}
\end{figure}

Discussion on the domains in which conversational agents have a relevant presence from a scientific perspective is diverse. Some domain-specific secondary studies offer a detailed dissertation of a specific area (e.g., healthcare \cite{Kocaballi2019-my}), while others report a domain taxonomy for a specific research context (e.g., business applications \cite{De_Barcelos_Silva2020-gw}). 
Based on the surveyed studies, we suggest the categorization covered in Figure \ref{fig:f2-sum} which allows us to identify 
6 general domains or research areas: 

\begin{itemize}
    \item \textbf{Daily life.} The most common domain among research is the support of daily life activities. It is reported as the most frequent domain by de Barcelos et al. \cite{De_Barcelos_Silva2020-gw}, Janssen et al. \cite{Janssen2020-fp}, Knote et al. \cite{Knote2018-dx} \add{and Guerino et al. \cite{Guerino2020-vc}}. The core of this area of application is the use and integration of conversational agents to achieve user's personal goals in daily life activities. In addition to general-purpose chatbots like \ed{crowd-powered Q\&A agents serving as smart search engines \cite{Savenkov2016-a}}, this domain includes different domain-specific application areas like tourism \cite{Niculescu2003-ee}, restaurants and food \cite{Kim2014-ht}, games \cite{Kobayashi2015-le}, sports \cite{Soros2013-pd} and smart-home assistance \cite{Fernando2016-ff}. 
    
    \item \textbf{Commerce.} Activities related to commercial activities, transactions and support are highlighted as the second most present area of application. While being reported by far as the most frequent domain by Bavaresco et al. \cite{Bavaresco2020-md}, it is also the second one reported by Janssen et al. \cite{Janssen2020-fp}. This domain includes e-customer service support chatbots \cite{Gnewuch2017-qm} and e-commerce assistance \cite{Vegesna2018-ak} like on-line shopping assistance \cite{Yan2017-jw}.
    \item \textbf{Business support.} Conversational agents in this domain are designed as software-based tools for employee support to internal business processes. The main goal is to provide an efficient, intuitive interface for traditional business processes to achieve a semi-automatic performance of such processes. These processes include financial tasks \cite{Okuda2018-co}, negotiation \cite{Zhao2018-uu} and team working or work support \cite{Bittner2019-oj}. 
    
    \item \textbf{Technical infrastructure.} Bavaresco et al. \cite{Bavaresco2020-md}, de Barcelos et al. \cite{De_Barcelos_Silva2020-gw} \add{and Guerino et al. \cite{Guerino2020-vc}} report technical management and technical support chatbots in a sub-set of dedicated sub-domains providing access and assistance management to complex technological infrastructures and technical processes. Chatbots reported by literature that can be matched into this category are typically related to complex technical infrastructure management and \ed{technical user assistance}. This domain covers from end-user technical support \cite{Subramaniam2018-ad} to advanced, semi-automatic management of smart industrial infrastructures \cite{Hauswald2016-lz}, including telecommunication systems like autonomous call-centres \cite{Mukherjee2014-js} and tools for bridging the gap in terms of usability between users and software products or tools \cite{Hsieh2014-bd}.
    
    \item \textbf{Healthcare.} Medical or health-related discussion is a matter of dedicated research in several secondary studies in our data-set \cite{Kocaballi2019-my, Milne-Ives2020-zl, Safi2020-cy}. However, the healthcare domain is also significantly present in domain-independent literature reviews \add{\cite{De_Barcelos_Silva2020-gw, Guerino2020-vc}}. Most common areas of application in this domain include support tools for healthcare professionals in medical activities like medication prescription \cite{Ahmad2018-at} or vital sign control and monitoring \cite{Santos2016-fd}, and support tools for patient assistance like therapy management \cite{Allen2018-th}.
    
    \item \textbf{Education.} Similarly to healthcare, dedicated research in the education field is also found in our data-set \cite{Hobert2019-gl, Perez2020-xt}, while it is also highlighted as a relevant area in domain-independent research \add{\cite{De_Barcelos_Silva2020-gw, Guerino2020-vc}}. Education conversational agents can be found as e-learning tools for student assistance in the learning process \cite{Todorov2016-ae} and as automated tutoring agents \cite{Suleman2016-pj}.
\end{itemize}

\subsection{Goals (F3)}
\label{subsec:f3}

\begin{figure}[h]
  \centering
  \includegraphics[width=0.89\linewidth]{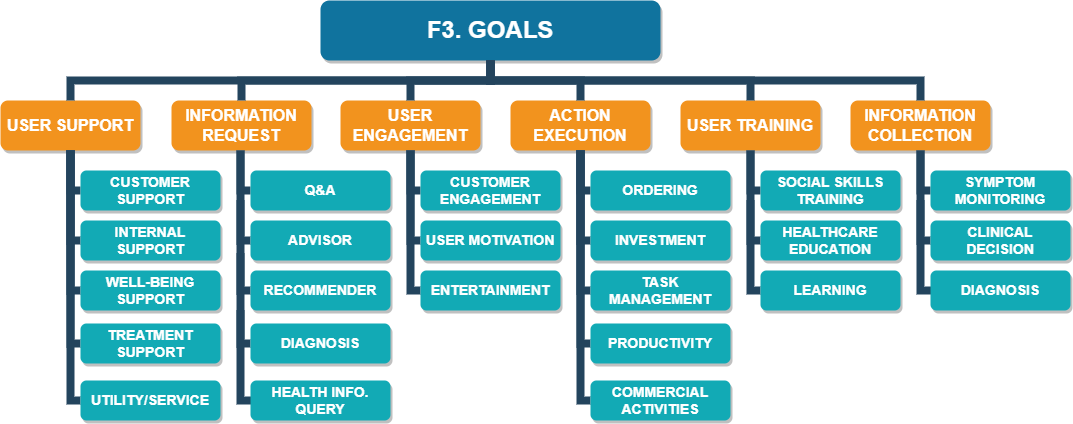}
  \caption{Aggregated results of conversational agents' goals in related literature}
  \label{fig:f3-sum}
\end{figure}

Figure \ref{fig:f3-sum} synthesizes and reports a general categorization of 6 major goals fitting the goals reported by the literature in the conversational agents field:  

\begin{itemize}
    \item \textbf{User support.} The most common goal in the field is to extend software systems with a support tool by integrating human-like communication features into existing activities and processes to enhance user experience. 
    This goal is present in many contexts, e.g: in the commerce domain, as customer and service support tools \cite{Bittner2019-oj}; in the business domain, as semi-automated interfaces of utility processes and for internal support \cite{Cui2017-pn}; in the healthcare domain, as therapy assistance like well-being psychotherapy \cite{Allen2018-th} and treatment support \ed{\cite{Morbini2012}}. 
    
    \item \textbf{Information request.} Conversational agents designed as commodity query engines giving easy-to-use access to knowledge databases (either generic or domain-specific) compose the second most common goal category. Conversational agents for information request include Q\&A chatbots \cite{Dharo2015-kb} and utility experts for complex domains like autonomous diagnosis or access to health-related information \cite{Philip2017-tw}. They also include daily-use, domain-specific utilities like restaurant recommendation \cite{Kim2014-ht} and tourist-experience advisor \cite{Niculescu2014-kl}. The information is either directly queried in a set of data sources and reported to the end-user, or processed through information aggregation algorithms to provide personalized answers based on auxiliary data (e.g., user information, context data).
    
    \item \textbf{User engagement.} Either as a primary or as a secondary goal, most conversational agents' research cover at some point user engagement with the conversational process itself or with the activity or process the agent is supporting. This engagement applies to a variety of domains, including customer engagement as in e-commerce \ed{\cite{qiu-etal-2018-transfer}}, motivation of the users for a specific activity like patient engagement in physical activity \cite{Kocielnik2018-tr}, and user engagement in entertainment-focused conversational agents \cite{cleverbot}. Research focusing on this goal is primarily oriented to key HCI factors in terms of user interaction and the agent's communication style.
    
    \item \textbf{Action execution.} Conversational agents are also used as text or speech interfaces giving access to a set of features from integrated or third-party software systems through a semi-autonomous communicative protocol. This goal is generally observed as a cross-domain objective, including online sale ordering \cite{Jusoh2018-mh}, team collaboration task management \cite{Toxtli2018-ot} and even domain-specific expert assistance like healthcare task productivity support \cite{lpl}. The common feature among these examples is to perform actions and features using natural language (either text or speech) as the interaction bridge between users and the system.
    
    \item \textbf{User training.} A wide community of chatbot research focuses on evaluating, training and improving user skills in a domain-specific area. This goal is especially prominent in the education field \cite{Perez2020-xt}, where learning is identified as a primary goal. But user training is not restricted to education: healthcare and medical research focus on training in the means of social skills training for mental spectrum disorders \cite{Tanaka2017-ay}, training of psychotherapies like mindfulness \cite{Hudlicka2013} and even using virtual patients as a learning tool \cite{Isaza-Restrepo2018-fl}. 
    
    \item \textbf{Information collection.} Collecting and processing information from users (either through the natural language interaction or, in some cases, through contextual data) is also the main objective in some research scenarios. This goal is mainly observed in healthcare-related activities like symptom monitoring \cite{Rhee2014-am}, diagnosis \cite{Havik2019-wf} and even advanced clinical decision or triage support \cite{Spanig2019-es}. Implicit and explicit data are combined to build and explore complex knowledge about user profiling to enhance and support other processes.
\end{itemize}

\subsection{Findings of SQ1}

Below we summarize the main insights that can be extracted from the features of SQ1.






\noindentparagraph{\textbf{Unclear terminology, but a predominant use of conversational agents and chatbots.}}
Research literature in the field is not built upon a clear taxonomy of terms and synonyms for the term variants used to refer to dialogue software-based systems. \textit{Conversational agents} and \textit{chatbots} are reported as the most common terms, and they are used in a majority of contexts as synonyms \add{\cite{Rheu2021-fq, Perez2020-xt, Nuruzzaman2018-iv,Dsouza2019-px}}. 
\add{\noindentparagraph{\textbf{Limited pervasiveness in specific research areas.}} Despite its increasing popularity in recent years, the adoption of conversational agents in some target-specific (e.g., elderly \cite{Costa_2018}) and domain-specific (e.g., professional healthcare \cite{Santos2016-fd}) areas is still limited to academic and research purposes \cite{De_Barcelos_Silva2020-gw}. Additionally, there is a lack of dedicated research from the organizational, stakeholder and ethical perspective in the field of conversational agents \cite{Syvanen2020-nb}.}

\noindentparagraph{\textbf{Domain/goal alignment.}}
The proposed taxonomies allow to identify some synergies between the identified goals and domains. User training is exclusively covered by education (e.g., student training \add{\cite{Todorov2016-ae}}) and healthcare (e.g., patient rehabilitation \add{\cite{Allen2018-th}}). Information collection is mainly related to healthcare (e.g., symptom monitoring \add{\cite{Santos2016-fd}}). Action execution is predominant in commerce (e.g., shopping \add{\cite{Yan2017-jw}}) and business support (e.g., scheduling \add{\cite{Bittner2019-oj}}). User support, information request and user engagement are majorly reported as cross-domain goals \add{\cite{Bavaresco2020-md, Janssen2020-fp, Nuruzzaman2018-iv}}.

\noindentparagraph{\textbf{User-centred goals.}}
User support, user engagement and user training require the active involvement of the user in the design and validation of the system \add{\cite{Kocaballi2019-my, Milne-Ives2020-zl}}. Consequently, the need for dedicated research in terms of user experience, user engagement and HCI is reinforced by these findings.

\noindentparagraph{\textbf{User adherence beyond proof-of-concept tools.}}
User engagement to conversational agents integrated as frequently used tools is a major challenge in some areas \add{\cite{Bavaresco2020-md, Kocaballi2019-my, Janssen2020-fp}}. Specific domains like healthcare, education and technical infrastructure seem to benefit from the user-oriented features of their disciplines. Consequently, the focus on the design of conversational agents as user-centred tools is key for a successful adherence in specific domains.

    
    
    

%% file: content/3_2_hci_features.tex
\section{Impact of HCI features (SQ2)}
\label{sec:sq2}

\subsection{HCI features (F4)}
\label{subsec:f4}

The analysis and integration of HCI features in conversational agents is subject of dedicated discussion as the main research goal in some literature reviews (e.g., Van Pinxteren et al. \cite{VanPixteren2020-zm}, Feine et al. \cite{Feine2019-vc}). In these studies, the authors propose their own taxonomy of what they refer to as \textit{social cues} or \textit{communicative behaviours}, respectively. On the other hand, some literature reviews (e.g., Rheu et al. \cite{Rheu2021-fq}, Janssen et al. \cite{Janssen2020-fp}) provide some valuable insights in the form of a narrative discussion as an enumeration of features and HCI-related conclusions supported by examples.

In our research, we focus on the concept of HCI features or \textit{social cues} as defined by Feine et al. \cite{Feine2019-vc}, where these are defined as design features that trigger a social reaction of the user towards the experience of use of the conversational agent. Under this definition, they propose a flat taxonomy of 42 social cues under a 4-class categorization based on communication features:

\begin{itemize}
    \item \textbf{Verbal} cues, including features about the content of the agents' responses (e.g., small talk) and the communicational style (e.g., dialect).
    \item \textbf{Visual} cues, including kinesics (e.g., movement), proxemics (e.g., background and conversational distance), appearance (e.g., physical attributes of the agent), and computer-mediated communication techniques (e.g., use of buttons).
    \item \textbf{Auditory} cues, including voice qualities (e.g., voice pitch) and vocalizations (e.g., laugh).
    \item \textbf{Invisible} cues, including chronemics (e.g., response time) and haptics (e.g., tactile touch).
\end{itemize}

On the other hand, Van Pinxteren et al. \cite{VanPixteren2020-zm} propose a two-dimension taxonomy of communicational features. The first dimension is a classification based on the type or \textit{modality} of such features, similar to the classification proposed by Feine et al. \cite{Feine2019-vc} but less detailed and more limited in the type granularity. The second dimension is based on the intention, the purpose or the \textit{footing} of these features, for which they propose a 3-class categorization:

\begin{itemize}
    \item \textbf{Human similarity}, which encompass the degree to which the user perceives that a conversational agent is relatable as a human being. These features are user-independent. They apply to general HCI rules and they are not submitted to context or individual differences (e.g., congruency of body gestures, physical appearance).
    \item \textbf{Individual similarity}, which encompass the degree to which users relate as individuals with the agent. These features are user-dependent. They apply to the capability of a conversational agent to match and adapt to the individual features of the user's needs, preferences, goals and personal communication styles (e.g., personality, communication style).
    \item \textbf{Responsiveness}, which encompass design factors that determine how a conversational agent reacts to a specific user interaction. These features can either be user-independent (e.g., politeness, small talk) or user-dependent (e.g., empathy, social praise).
\end{itemize}

All in all, while Van Pinxteren et al. \cite{VanPixteren2020-zm} report a deeper approach in terms of verbal features type classification, Feine et al. \cite{Feine2019-vc} offer a broader analysis scope by introducing the footing dimension as a criterion for structuring and describing communicational features. Therefore, we conclude as a task of significant value to cross-reference both taxonomies to compensate the strengths and weaknesses of both approaches and to integrate the knowledge and dissertation they provide.

\begin{figure}[h]
  \centering
  \includegraphics[width=0.95\linewidth]{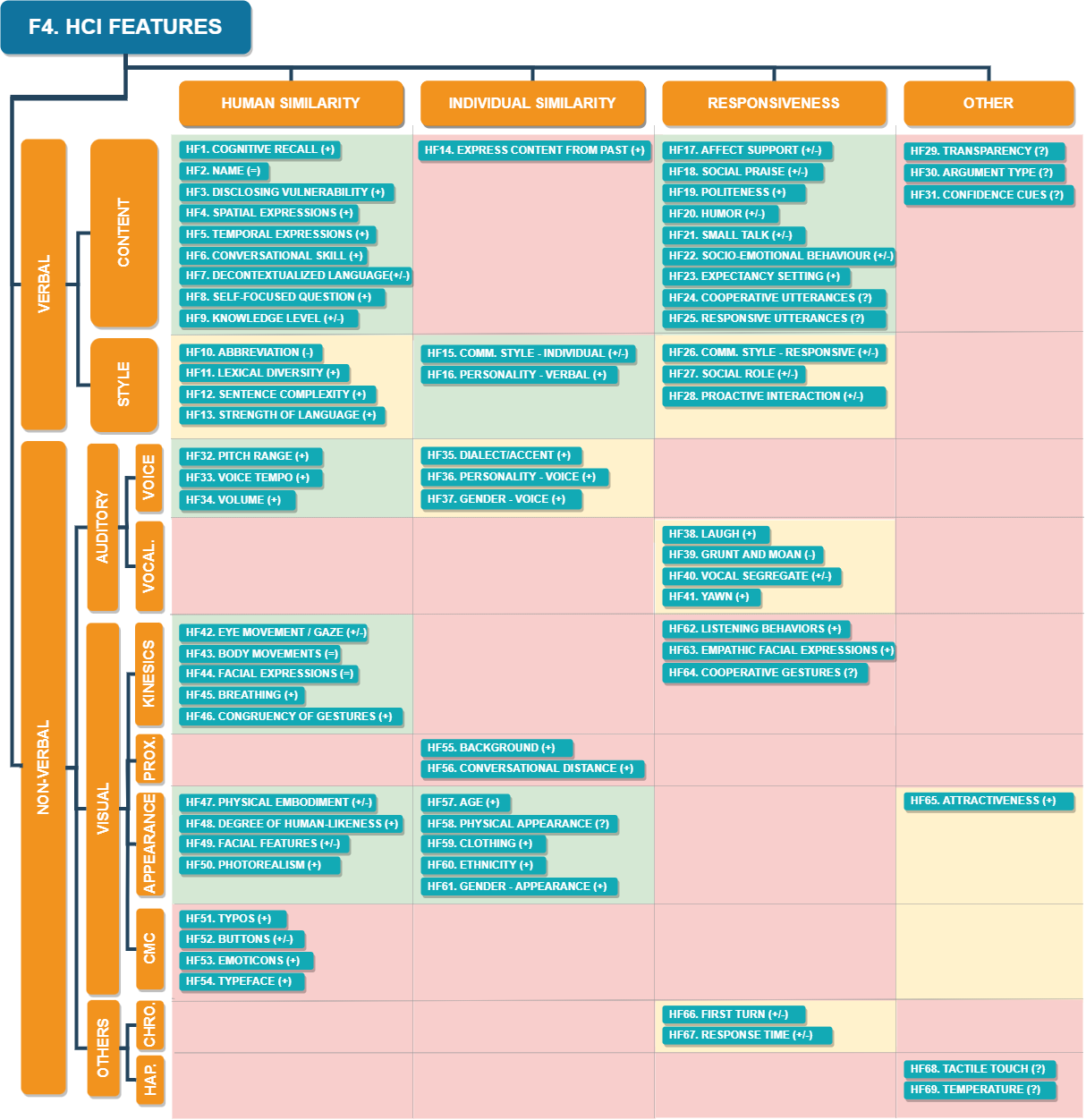}
  \caption{Aggregated results of HCI features for the design of conversational agents in related literature, including impact evaluation and a heat-map of the frequency of use}
  \label{fig:f4-sum}
\end{figure}

In addition to these taxonomies, Rheu et al. \cite{Rheu2021-fq} use a narrative approach to discuss major themes related to perceived trust in the use of conversational agents, including social intelligence of the agent, voice characteristics and communication style, anthropomorphic look of the CA, non-verbal communication, and performance quality. 
Among the main conclusions, they emphasize human-like similarity including verbal communication and expressions as well as non-verbal features, although they stress out the impact and the effect of the latter, especially in embodied conversational agents. Additionally, they also discuss the importance of individual similarity in terms of personification of the agent that suits the users (e.g., formality, ethnicity, communication metrics). 
Janssen et al. \cite{Janssen2020-fp} depict a chatbot taxonomy for design dimensions from three different perspectives: intelligence, context and interaction. Regarding the latter, they mention a subset of 7 design dimensions which can be mapped to either one of the two complete taxonomies previously mentioned. Some examples include interface personification (i.e., disembodied vs. embodied) and user assistance design (i.e., reactive communication vs. proactive communication). 
\add{Zierau et al. \cite{Zierau_2020} present a categorization of HCI variables in the form of aggregated dimensions based on the taxonomy presented by Feine et al. \cite{Feine2019-vc}. They extend this contribution by analyzing and coding the impact dimension of these variables with respect to perceptual qualities. Some examples include the likeability of the agent \cite{Chin_2019}, it's perceived humanness \cite{Candello_2017} and user satisfaction \cite{Chaves_2018}.}

After aligning the taxonomies defined by Van Pinxteren et al. \cite{VanPixteren2020-zm} and Feine et al. \cite{Feine2019-vc} with the aggregated knowledge obtained from the rest of studies, we harmonized the structure and definitions of these HCI features. The results of this process are summarized in Figure \ref{fig:f4-sum}, and the contributions of such results can be summarized as follows:

\begin{itemize}
    \item A two-dimension taxonomy of \textbf{69 HCI features}. The first dimension is the modality or type taxonomy of HCI feature as proposed by Feine et al. \cite{Feine2019-vc}, while the second dimension is the footing as defined by Van Pinxteren et al. \cite{VanPixteren2020-zm}. 
    \item A synthesis on the \textbf{impact evaluation} regarding how each HCI feature affects the communicational process between the user and the agent using a 5-value scale: 37 with positive impact (+); 3 with negative impact (-); 17 with mixed conclusions, typically depending on the context or scenario of use (+/-); 3 with neutral or undetectable impact (=); and 10 with no clear discussion or with missing empirical evaluation conclusions (?).
    \item A \textbf{heat-map} of the use and frequency of these HCI features based on their mentions in the secondary studies included in this feature, which allows identifying specific categories from the cross-referenced taxonomy with different levels of maturity and relevance. \add{HCI categories coloured in red are evaluated as rare or missing, given that: (1) they are covered by either one or none of the surveyed papers; and (2) the majority of features include only one example for each reported feature. Categories in yellow are evaluated as uncommon, given that: (1) they are only mentioned by one of the surveyed papers; and (2) the majority of reported features are supported by multiple examples. Finally, categories in green are evaluated as common, given that they are mentioned and exemplified by several of the surveyed papers. The resulted heat-map allows a visually assisted representation of the state-of-the-art of HCI features in the field of conversational agents in terms of their maturity and pervasiveness.}
\end{itemize}

\subsection{Findings of SQ2}

In this section, we summarize the main insights that can be extracted from the results of SQ2.

\noindentparagraph{\textbf{Research focuses on verbal content features for human similarity and responsiveness.}}
As reported during the analysis of the design and technical specifications (F6), the building of a consistent and adequate knowledge base for the appropriate natural understanding and dialogue management performance is essential for a successful user experience \add{\cite{Adamopoulou2020-yj}}. Moreover, adaptation in terms of contextualized and personalized verbal content \add{\cite{Bavaresco2020-md}} is one of the main sources of adaptation reported in recent literature (F7). All in all, research is clearly positioned on the importance of exploring and analysing the impact of verbal content HCI features. When focusing on verbal communicational style, reported HCI features are less frequent, and literature generally reports mixed conclusions with respect to their impact, especially from the responsiveness dimension of the agent (e.g., a proactive character is not always perceived as a positive feature \add{\cite{Liao-2016}}).

\noindentparagraph{\textbf{Absence of communicational features towards individual similarity.}} The only reference is the capability of the agent to express content from past conversations with the users \add{\cite{Hastie_2016}}, which on the other hand is reported to be undoubtedly positive in terms of trust and perceived naturalness of the agent \add{\cite{Feine2019-vc}}. As reported in the context evaluation (F7), the potential and benefits of integrating contextual and historical data not only in terms of past interactions \add{\cite{Subramaniam2018-ad}} but also as extended knowledge sources from the user environment \add{\cite{Kocaballi2019-my}} is a key feature to be further investigated to enhance the user experience and improve user engagement.  \add{Beyond purely verbal content features, we identify: (1) a lack of structured knowledge regarding the influence between the conversational agent communicational style and engagement with the agent \cite{Zierau_2020}; and (2) a research gap in terms of non-verbal communicative behaviours \cite{VanPixteren2020-zm} and potential adaptivity features for personalized conversational agents \cite{Guerino2020-vc}.} Consequently, there is room for research focused on content adaptation and personalization towards individual similarity based on users' unique needs.

\noindentparagraph{\textbf{Appearance features are more important towards individual similarity rather than human similarity.}} 
Individual similarity in terms of visual proxemics and appearance are generally reported to have a positive impact on user experience and user engagement \add{\cite{Benlian_2019, Yuksel_2017}}. On the other hand, human similarity in kinesics and appearance report mixed or context-dependent results in terms of impact, as reproducing some human behaviours (e.g., eye monitoring, facial features) can have a negative impact in some contexts in terms of anxiety \add{\cite{Rickenberg_2000}} and performance \add{\cite{Salem2011}}. 

\noindentparagraph{\textbf{Computer-mediated communication can have a positive impact on user experience.}} These communicational features allow complementing communication messages through alternatives to natural language (e.g., buttons \add{\cite{Diederich_2019}}, emoticons \add{\cite{Park_2015}}). Literature proves that these techniques not only help to achieve human similarity \add{\cite{Guerino2020-vc}} but also can be a useful approach for other considerations like to reduce typing effort and avoid human errors \add{\cite{Jain2018-kk}}, which consequently have a significant impact on relevant quality characteristics of the field like the robustness of unexpected input (F9).\\

    

%% file: content/3_3_technical_methods.tex
\section{Design and implementation (SQ3)}
\label{sec:sq3}

\subsection{Design dimensions (F5)}
\label{subsec:f5}

\begin{figure}[h]
  \centering
  \includegraphics[width=0.88\linewidth]{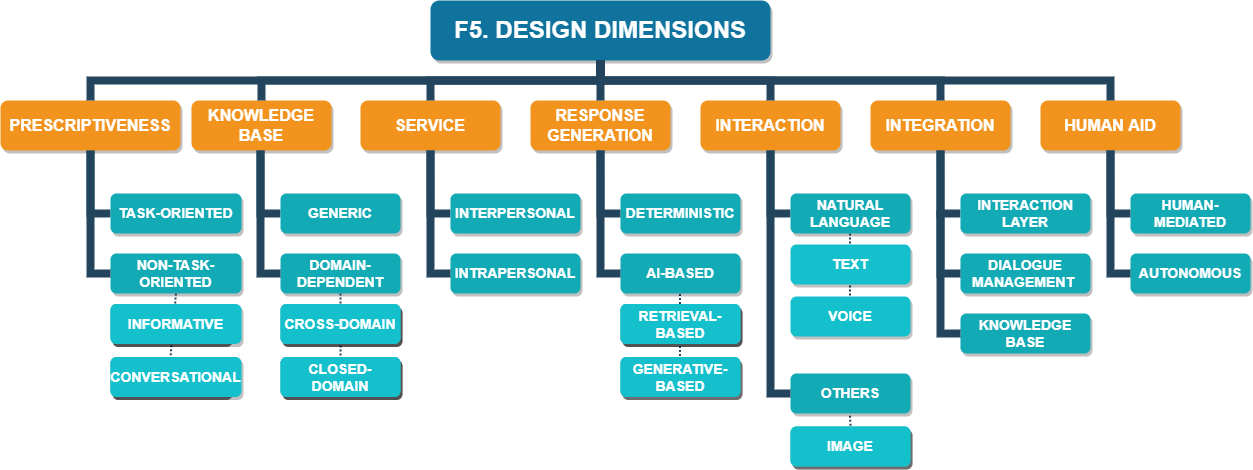}
  \caption{Aggregated results of conversational agents' design dimensions in related literature}
  \label{fig:f5-sum}
\end{figure}

Research literature reports high-level design dimensions for conversational agents in the form of categories and subcategories proposing a taxonomy based on design specifications. By synthesizing this knowledge, in Figure \ref{fig:f5-sum} we report a summary of \ed{7} design dimensions based on the taxonomy proposed by Adamopoulou and Moussiades \cite{Adamopoulou2020-yj} and the aggregated knowledge from the research literature included in our study. We describe these dimensions as follows: 

\begin{itemize}
    \item \textbf{Prescriptiveness.} Hussain et al. \cite{Hussain2019-lj} present two major design categories for classifying conversational agents based on their goals: \textit{task-oriented} and \textit{non-task-oriented}. \textit{Task-oriented} agents are defined as short-conversation agents designed to execute a particular action from a known sub-set of pre-configured tasks triggered by the conversational process. An online shopping assistant for solving order-related questions \cite{Yan2017-jw} is an example of a \textit{task-oriented} agent. \textit{Non-task-oriented} agents aim to simulate a human-conversational process without a specific task as the main goal of the interaction with the agent. Leisure or entertainment agents like Cleverbot \cite{cleverbot} fall into this category. With respect to \textit{non-task-oriented} conversational agents, Nuruzzaman and Hussain \cite{Nuruzzaman2018-iv} differentiate between \textit{conversational} and \textit{informative} agents. While \textit{conversational} agents match the definition of \textit{non-task-oriented} by Hussain et al. \cite{Hussain2019-lj}, \textit{informative} agents are defined as a type of \textit{non-task-oriented} agents which do not pursue a specific activity or task to be executed, but the interaction and the conversational process has the purpose of requesting information. Q\&A and service support chatbots fall into this category \cite{Fernando2016-ff,Kim2014-ht,Niculescu2014-kl}. \textit{Task-oriented} agents typically imply either the agent is integrated as a tool or sub-module of another software system providing a set of features to the user, or the agent requires integration with third-party software services to perform these actions. 
    
    \item \textbf{Knowledge base.} A major design key factor is the scope of the knowledge base that conversational agents support in their conversations. This design feature has an impact on the data sources (in terms of number, variety and complexity) required to build the knowledge base. Adamopoulou and Moussiades \cite{Adamopoulou2020-yj} differentiate between \textit{generic} and \textit{domain-dependent} knowledge bases. \textit{Generic} knowledge bases offer information and conversation topics from any domain. An example of this category is a self-adaptive agent designed to learn and adapt to new contexts and topics based on user interaction \cite{Huang2018-cb}. On the other hand, \textit{domain-dependent} agents include open or \textit{cross-domain} (i.e., integrating multiple or several knowledge bases and domain data sources \cite{Savenkov2016-a}) and \textit{closed-domain} (i.e., focusing on a single, expert knowledge base \cite{Kucherbaev2018-bh}), which are the two main categories covered by Hussain et al. \cite{Hussain2019-lj} and Nuruzzaman and Hussain \cite{Nuruzzaman2018-iv}. \textit{Generic} and \textit{cross-domain} chatbots typically require auxiliary NLP techniques to process and contextualize user input into a specific topic or domain, like co-reference resolution techniques \cite{Adamopoulou2020-yj}. This is a consequence of the broader scope of the conversation and user intents and entities. On the other hand, \textit{closed-domain} chatbots are more likely to be tightened to a specific, well-known subset of intents and entities. 
    
    \item \textbf{Service.} The service dimension is defined as the type of relationship established between the chatbot and its users, based on the needs fulfilled by the agent. Nuruzzaman and Hussain \cite{Nuruzzaman2018-iv} differentiate between \textit{interpersonal} and \textit{intrapersonal} agents. \textit{Interpersonal} agents do not build a personal relationship with the user and are focused to provide a generic service based on a general user categorization. An example of this category is a conversational agent for restaurants' recommendation and reservation \cite{Kim2014-ht}. On the other hand, \textit{intrapersonal} agents are focused on personal scenarios where the chatbot helps users fulfilling tasks in their personal life, and therefore there is a personification of the service based on user needs. For instance, smart home assistants for the elderly \cite{Fernando2016-ff} fall into this category. While \textit{interpersonal} agents are typically context-independent and user-independent, \textit{intrapersonal} agents are designed with user profiling and user configuration mechanisms to fit to the user's needs.
    
    \item \textbf{Response generation.} This design dimension relates to the mechanism used by the conversational agent to generate an adequate response based on the user input messages. Literature primarily differentiates two major categories of solutions: \textit{deterministic} mechanisms and \textit{\ed{ML-based}} mechanisms. \textit{Deterministic} algorithms process user input messages to extract some kind of structure, interpreted knowledge, and apply some kind of deterministic strategy to link this structured data to a specific output message or action. 
    On the other hand, more recent strategies are exploring the potential of \textit{\ed{ML-based}} strategies, which integrate the use of machine learning and DL models to process user input and build output messages based on the knowledge sources and training data. Mainly, there are two types of \textit{\ed{ML-based}} strategies: \textit{retrieval-based} and \textit{generative-based}. As defined by Adamopoulou and Moussiades \cite{Adamopoulou2020-yj}, \textit{retrieval-based} systems use ML/DL models and techniques to predict the most accurate response from a closed set of responses using an output ranked list of possible answers. On the other hand, \textit{generative-based} systems focus on using DL models to synthesize and build the reply to a specific user input, rather than selecting it from a closed data-set of responses. 
    
    \item \textbf{Interaction.} This dimension defines the communication mechanism used by the conversational agent to process user information and to generate responses to the user. 
    The core of related research is focused on the design and development of \textit{natural language} interfaces supported by some kind of natural language pre-processing pipeline or natural language understanding module (see Section \ref{subsec:f6}) to extract interpretable knowledge from natural language messages. To this end, conversational agents use \textit{text} interfaces, \textit{voice} recognition or a combination of both, potentially adapting the medium to the needs and the context of the user at each stage of the dialogue \cite{Rheu2021-fq}.
    Typically, agents integrating \textit{voice} interaction with their users introduce some kind of speech-to-text and text-to-speech or automatic speech recognition systems as a top layer of the conversational process to support both mechanisms. This is the case of well-known commercial voice assistants like Alexa or Cortana \cite{Matthew2018}. 
    In addition, some studies introduce complementary interaction mediums to overcome the limitations of natural language using alternative data formats. 
    Adamopoulou and Moussiades \cite{Adamopoulou2020-yj} report \textit{image} processing as a valuable mechanism for user interaction. Some agents introduce image recognition as a feature to support and extend the limitations of natural language communication, supported by other commercial, popular agents like Siri or XiaoIce \cite{Shum2018-xd}. \add{In addition to these interaction mechanisms, Pérez-Soler et al. \cite{Perez-Soler_2021} comment on advanced, optional technical factors which have a design impact on the aforementioned communication mediums, including multi-language support for voice and text interaction \cite{Adamopoulou2020-yj} and sentiment analysis of natural language input messages \cite{Folstad_2018}.}  
    
    \item \add{\textbf{Integration.} This dimension refers to the software ecosystem in which the conversational agent is designed, developed and deployed and the integration of software systems and services from a high-level design perspective. We conceive dialogue-based software systems as defined by Bhirud et al. \cite{Bhirud2019-ml}, which include: the \textit{interface} layer, the \textit{dialogue management} layer and the \textit{knowledge base} layer. Concerning the \textit{interface} layer, the use of popular distribution platforms like social networks (e.g., Facebook \cite{brixey-etal-2017-shihbot}) and other communication channels provide a user base and a technical infrastructure to easily deliver dialogue-based software solutions \cite{Lebeuf2018}. Concerning the \textit{dialogue management} layer, Pereira et al. \cite{Pereira_2018} compare and categorize a subset of software components for the development of conversational agents and natural language understanding modules, including libraries, frameworks, platforms and services (see Section \ref{subsec:f6}) which have a significant impact on development, testing and deployment stages (e.g., API invocations to third-party external services \cite{Zamanirad_2017}). Pereira et al. \cite{Pereira_2018} also comment on integrating third-party systems for either automatic or supervised dialogue management tracking. Finally, concerning the \textit{knowledge base} layer, integration with external databases and information systems allows benefitting from big data repositories, leveraging the required effort to maintain and update these systems \cite{Zumstein_2017}.}
    
    \item \textbf{Human aid.} Depicts the degree of autonomy in which the conversational agent can be handled, whether it is designed as a \textit{human-mediated} or an \textit{autonomous} agent. As depicted by Adamopoulou and Moussiades \cite{Adamopoulou2020-yj}, \textit{human-mediated} refers to agents which require from human computation at some point in the conversational process to be operated \cite{Kucherbaev2018-bh}. On the other hand, \textit{autonomous} agents are fully operated autonomously by users without human-assistance in the loop. While information processing speed might become critical in \textit{human-mediated} agents due to the need of integrating human-aided steps, some contexts of use like crowd-sourcing business services \cite{Huang2018-cb} require from human workers to adapt and extend the agent's knowledge base at runtime.
\end{itemize}

\subsection{Technical implementation specifications (F6)}
\label{subsec:f6}

\begin{figure}[h]
  \centering
  \includegraphics[width=0.95\linewidth]{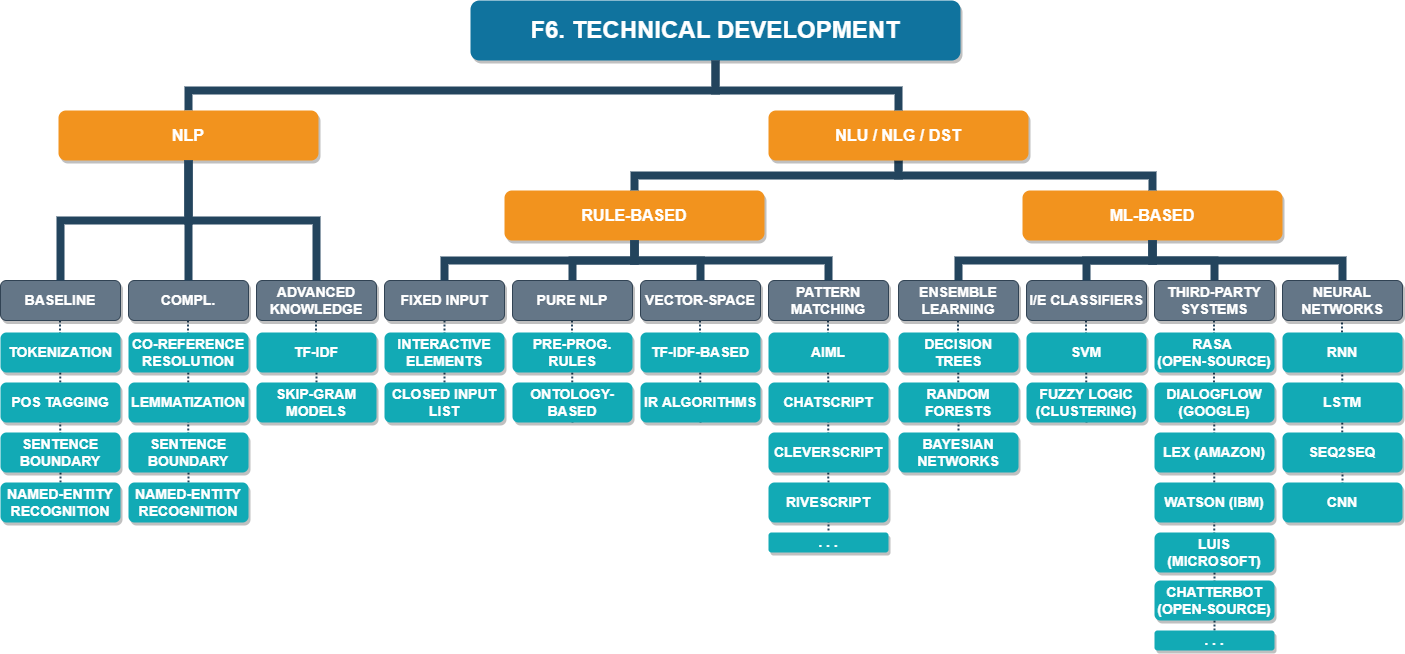}
  \caption{Aggregated results of methods, techniques and technologies in related literature}
  \label{fig:f6-sum}
\end{figure}

Analysis of the methods, techniques and technologies used for the implementation of conversational agents is one of the most frequent subjects of dedicated research in our study data-set. Some secondary studies integrate technical specifications as a minor goal of their research, providing a flat enumeration of techniques and technologies aligned with relevant examples \cite{Masche2018-sw}. Contrarily, we also found dedicated surveys for the analysis of technical aspects presenting taxonomies for technical development, including domain-specific \cite{Safi2020-cy} or cross-domain \cite{Janssen2020-fp}. 
Concerning natural language processing and knowledge interpretation, we identify four major topics involved in the implementation of conversational agents: natural language processing (NLP), natural language understanding (NLU), natural language generation (NLG) and dialogue stage tracking (DST). Figure \ref{fig:f6-sum} summarizes the methods, techniques and technologies reported by research literature.

Concerning NLP techniques, given that natural language is the essential core of the conversational process of the majority of chatbot interfaces, these are typically included as a top layer between the user input interface (e.g., chat interface, voice recognition) and the natural language interpretation module responsible for interpreting these messages. Enumerations of NLP techniques are present in research literature, but they are frequently presented as a simplified, preliminary stage \cite{Bhirud2019-ml} or even as a secondary task \cite{Abdul-Kader2015-jy}. We suggest classifying these techniques into three categories: \textit{baseline techniques}, \textit{complementary techniques} and \textit{advanced knowledge interpretation}.

\begin{itemize}
    \item \textbf{Baseline techniques.} Used commonly as preliminary tasks to any NLP pipeline, base techniques are used to structure, clean and assign basic grammatical annotations to the user input messages \cite{Bhirud2019-ml}. They are generally integrated with independence of the integration of a NLU module. Base techniques include tokenization, part-of-speech (POS) tagging, sentence boundary disambiguation and named entity recognition.
    \item \textbf{Complementary techniques.} As an extension of base techniques, complementary techniques are typically applied before advanced NLU techniques to obtain extended grammatical and semantic knowledge \cite{Masche2018-sw}. Complementary techniques include co-reference resolution, lemmatization, dependency parsing and semantic role labelling.
    \item \textbf{Advanced knowledge interpretation.} Advanced knowledge structures like the use of vector-space models are the bridge between pure NLP techniques and the integration of NLU modules. This advanced interpretation allows transforming text corpus data into a knowledge base for the conversational agent communicational process \cite{Maroengsit2019-by}. These techniques include vector-based representations like TF-IDF and skip-gram models like Word2Vec.
\end{itemize}

Concerning NLU, NLG and DST techniques, and as discussed in Section \ref{subsec:f5} (F5), secondary studies discussing the technical development of the NLU and NLG mechanisms agree on a general categorization of two main approaches: \textit{rule-based} approaches and \textit{\ed{ML-based}} approaches.

\begin{itemize}
    \item \textbf{Rule-based}. Based on the definition in Section \ref{subsec:f5}, we identify four technique categories: \textit{fixed input}, \textit{pure NLP}, \textit{vector-space} and \textit{pattern-matching}.
    \begin{itemize}
        \item \textit{Fixed input}. 
        Fixed input agents restrict the conversational process to a closed set of possible inputs to select at each stage of the dialogue. Despite being only explicitly mentioned by Safi et al. \cite{Safi2020-cy}, and even though pure fixed input does not require a natural language module, Janssen et al. \cite{Janssen2020-fp} present hybrid NLU approaches in which fixed input mechanisms at a specific stage of the dialogue are used to reduce typing effort and human errors by simplifying the user interaction. Some examples include interactive elements (e.g., buttons) \cite{Jain2018-kk} or multiple-choice selection of a dynamically updated list \cite{Bickmore2016-um}. 
        
        \item \textit{Pure NLP}. Some approaches using basic text understanding techniques integrate the use of base and complementary natural language processing techniques. These techniques are complemented alongside search algorithms for proper response selection through pre-programmed rules triggering specific answers when finding specific keywords \cite{Wang2018-zi} or advanced techniques like ontology-based matching through dependency parsing \cite{Gosh2018-bb}.
        
        \item \textit{Vector-space}. Advanced knowledge interpretation techniques can extend the NLP module to build interpreted domain knowledge through vector-space modelling techniques. These models can be used as mapping modules between user inputs and chatbot outputs in a variety of contexts. For instance, TF-IDF models can be used to build expert knowledge bases \cite{Lin2017-ll}, and vector-space models can also be exploited by more complex information retrieval (IR) algorithms for complex domain-specific queries \cite{Feng2006-sk}. 
        
        \item \textit{Pattern matching}. The most representative approach for rule-based agents is the development of a pattern matching approach through popular, standardized technologies like Artificial Intelligence Markup Language (AIML) \cite{aiml}, ChatScript \cite{chatscript}, Cleverscript \cite{cleverscript}, or RiveScript \cite{rivescript}. These are examples of open-source interpreted languages defining the syntax to build a set of templates to identify patterns of user input messages and their link with patterns of output responses. These syntaxes are designed using well-known data-interchange formats like XML (i.e., AIML), JSON (i.e., Cleverscript) or simplified custom formats (i.e., ChatScript, Rivescript). Despite their conceptual, syntactical and technical differences, all of them allow to define a knowledge base through a set of dialogue management rules. Through input processing, they apply pattern recognition techniques to find the most suitable template and proper response selection  for a specific user input. These approaches offer an easy-to-develop, efficient and effective method to manually define limited knowledge bases, perfectly suitable for closed-domain and specific-purpose chatbots where a limited conversation scope is expected. These contexts of use include examples like automated medical chatbots \cite{Rarhi2017-bm}, academic advisors for student assistance \cite{Latorre2015-hj} or a website-based chatbot for e-commerce support \cite{Gupta2015-bm}.
        
    \end{itemize}
    \item \textbf{\ed{ML-based}}. Based on the definition in Section \ref{subsec:f5}, we also identify four technique categories: \textit{ensemble learning}, \textit{I/E classifiers}, \textit{\ed{third-party systems}} and \textit{neural networks}.
    \begin{itemize}
        \item \textit{Ensemble learning}. In the context of conversational agents, ensemble learning models like decision trees or random forests are used for input text classification through a voting approach to determine the most suitable response through the tree paths definition and the user answers to the chatbot questions. These decision trees are built and can be dynamically adapted by adding new nodes (i.e., new user inquires) through new user interaction, and consequently tracing new paths and adapting to unknown conversational patterns. Ensemble learning models are used in the context of Q\&A chatbots for domain-specific contexts like education advisors \cite{Mondal2018-dd} and autonomous diagnostic agents \cite{Rida2018-az}. 
        Given the significant amount of conversational paths a conversation might undertake, they are complex to develop and maintain \cite{Chetty2015-vp}. However, in some contexts, they have proven to provide a higher accuracy that alternative methods like fuzzy logic approaches \cite{Bhirud2019-ml}.
        
        \item \textit{Intent/entity (I/E) classifier}. A standardized approach for developing NLU modules is the use of machine learning classifiers for automated, adaptive intent and entity recognition. In this domain, intents are defined as the user intentions, while entities are the topics these intents refer to. The use of pre-trained models through manual annotation and model evolution through runtime user interaction allows processing user input messages and to predict with a certain probability the most accurate response based on the intents and entities predicted from that message. I/E classification can be developed using a wide variety of classifiers, including Support Vector Machine (SVM) \cite{Mu2017-sk} and fuzzy logic (clustering) \cite{Chetty2015-vp}.
        
        \item \textit{\ed{Third-party systems}}. Given the increasing growth of conversational agents in a wide variety of domains (F1), several \ed{third-party} solutions for their development, deployment and maintenance have emerged in recent years. \ed{Using as reference the categorization presented by Pereira et al. \cite{Pereira_2018}, these third-party systems include libraries, frameworks, platforms and services designed as feature providers which cover from designing and training NLU modules to develop, deploy and operate chatbots in an easy-to-use, decoupled and scalable environment}. As reported by Safi et al. \cite{Safi2020-cy}, these solutions typically allow the combination of machine learning (i.e., I/E classifiers) and rule-based approaches for NLU tasks. Some examples include commercial solutions like Google Dialogflow\footnote{https://cloud.google.com/dialogflow/docs} \ed{(platform)}, Amazon Lex\footnote{https://docs.aws.amazon.com/lexv2/latest/dg/what-is.html} \ed{(platform)}, IBM Watson\footnote{https://www.ibm.com/watson} \ed{(platform)} and Microsoft Luis\footnote{https://azure.microsoft.com/en-us/services/cognitive-services/language-understanding-intelligent-service/} \ed{(service)}. Despite being proprietary software solutions, all of them include some kind of free-tier service (e.g., limited by the number of features or the amount of data that can be processed). Among open-source solutions, the most popular is Rasa \footnote{https://rasa.com/} \ed{(framework)}, which has become a technical reference in the conversational AI community. \add{Finally, some dialogue engines like Chatterbot\footnote{https://github.com/gunthercox/ChatterBot} (library) provide access to utility NLU modules for training, testing and generating conversational agents, which despite requiring logic from a specific programming language they generally offer higher degrees of customization and integration  \cite{Perez-Soler_2021}.}
        
        \item \textit{Neural networks}. Artificial intelligence neural networks are the latest technical contribution to the conversational agents' field. As described by Adamopoulou and Moussiades \cite{Adamopoulou2020-yj}, user input messages are transformed into vector representations through word embedding techniques \cite{Mikolov2013-cc}. These vectors are used as input features of the neural networks, which can be used to either predict (i.e., \textit{retrieval-based}) or create (i.e., \textit{generative-based}) the response.  
        A special type of neural networks has become more popular among conversational agent developers due to its capability of integrating contextual knowledge with respect to previous conversations. Recurrent Neural Networks (RNNs) define an architectural loop in which output knowledge data is fed as input features to the network cells continuously. Among RNNs, Long Short-Term Memory neural networks (LSTM) allow conversational agents to differentiate between long and short-term memory. Long-term memory refers to the general knowledge data of the model for conversation prediction and generation, while short-term memory refers to knowledge data that is only valid and relevant for a specific time window in which context is defined by recent interaction between the user and the agent \cite{Maroengsit2019-by}. This time window, which defines how cells behave and how they manage its contextual knowledge, can be configured according to the needs and nature of the agent.
        RNNs and LSTM are frequently used in \textit{generic} or \textit{open-domain} agents. Some examples include a highly accurate chatbot of frequently asked questions for customer service based on a \textit{retrieval-based} approach \cite{Muangkammuen2018-is} and a proposal of a \textit{generative-based}, knowledge-grounded chatbot model \cite{Kim2020-kk}.
        When focusing on \textit{generative-based} approaches, the use of Sequence-to-Sequence (Seq2Seq \cite{Sutskever2014-vl}) models is becoming predominant. Seq2Seq models use LSTM neural networks as layers to map the user input sequence into a target sequence, which is the generated response of the agent. As representative examples of context-aware and \textit{generative-based} conversational agents, we highlight a generic emotionally aware chatbot for user engagement through open-conversation \cite{Zhou2018-hz} and an open-domain question-answering chatbot for user support \cite{Qiu2017-lw}. 
        Complementary, some approaches use convolutional neural networks (CNN) not only for intent/entity classification \cite{Rai2018-rs}, but also in the context of image processing \cite{Shum2018-xd} as a complementary interaction mechanism with users for an extension of the knowledge base of the conversational agent.
    \end{itemize}
\end{itemize}

\subsection{Context integration techniques (F7)}
\label{subsec:f7}

Integrating contextual data is reported as a key feature and as a future research challenge at some point by most of the secondary studies in our data-set. While being limited, research literature discussing context integration focus on how conversational agents integrate context into their systems to provide a personalized, self-adaptive and context-aware user experience as defined by Bavaresco et al. \cite{Bavaresco2020-md} and Janssen et al. \cite{Janssen2020-fp}. 

Deep learning strategies like RNNs and LSTM introduce a context-adaptation layer based on a specific conversation-related context through recent user messages, which allows topic-centred analysis and runtime feedback to increase personalization and accuracy. But context integration strategies are manifold. Knote et al. \cite{Knote2018-dx} define context data as the knowledge that can be extracted from physical (e.g., location, temperature, humidity) and logical (e.g., calendar entries, application's portfolio) environments. \add{However, more relaxed definitions of context include past user-agent interactions and historical dialogue data \cite{rasa_2021}.} According to the feature extraction analysis from secondary studies, there is no formal analysis or categorization on how to integrate context into conversational agents. Instead, we find a summary enumeration of general techniques or specific examples. 

Bavaresco et al. \cite{Bavaresco2020-md} define three focused research question concerning self-learning, personalization and \textit{generative-based} methods in conversational agents which they address through a narrative approach commenting examples of their primary studies data-set. These examples are described emphasizing the data sources of contextual knowledge and the mechanisms used for context integration. Some examples include user personalization features \add{or \textit{user preferences}} and user's past interactions with the agent for \ed{\textit{adaptive knowledge bases}} to domain-specific queries \cite{Subramaniam2018-ad}, \ed{\textit{user feedback}} regarding the users' \ed{\textit{runtime feedback} and} the agents' responses to re-train the generative models \cite{Xue2018-ky}, and \ed{\textit{user historical data}} (e.g., user's purchase in an e-commerce website) from the software system in which the agent is integrated or from third-party services for \ed{\textit{recommender} tasks} \cite{Sapna2019-ca}. Kocaballi et al. \cite{Kocaballi2019-my} present an exhaustive enumeration of examples related to a list of personalized contents and context-adaptation purposes or goals. For the former, some examples include progress towards the goals set and communication style adaptation, as well as \add{customized \textit{responses} through \textit{content/style adaptation} like personalized} reminders, warnings and alerts sent by the agent to this end \cite{Sillice2018-mf}, and multimedia adaptation like customized activity graphs based on user activities as well as monitoring questions on these activities \cite{Kocielnik2018-tr}. For the latter, some examples include improving \ed{\text{user engagement}} and delivering \ed{\textit{adaptive training} for mental health activities} \cite{Hudlicka2013}, \ed{\textit{adaptive support}} tasks like analysis on self-reported symptoms of depression \cite{Inkster2018-ss}, \add{improve \textit{dialogue quality} for a better user experience \cite{Inkster2018-ss}, and provide \textit{timely feedback} through consultant monitoring \cite{Harper2008-nm}.}

Knote et al. \cite{Knote2018-dx} enumerate a few examples of context changes reactions, focusing on the mechanism and the object of these reactions or changes on assisting users on a specific task \add{or \textit{enacted actions}} like semi-automatic purchases \cite{Venkatesh2017-ah} 
. \add{Advanced approaches in complex systems introduce context-aware strategies for the adaptation of the software ecosystem beyond the conversational agent behavior itself, including \textit{smart objects} for smart home assisted adaptation\cite{Fernando2016-ff} and \textit{third-party services} for adaptive interoperability \cite{Stoeckli_2018}.}
Finally, Janssen et al. \cite{Janssen2020-fp} simply limit to report that \add{personalized agents with \textit{adaptive self} capabilities in terms of runtime communicational style and behavior personalization} are a significant minority in the data-set of chatbots they cover, with a significant presence of short-term relationship chatbots which do not consider context data. Despite being a recent contribution, they argue that this is a direct consequence of the rather complex design properties and technical specifications involved in these dimensions.

\begin{figure}[h]
  \centering
  \includegraphics[width=0.70\linewidth]{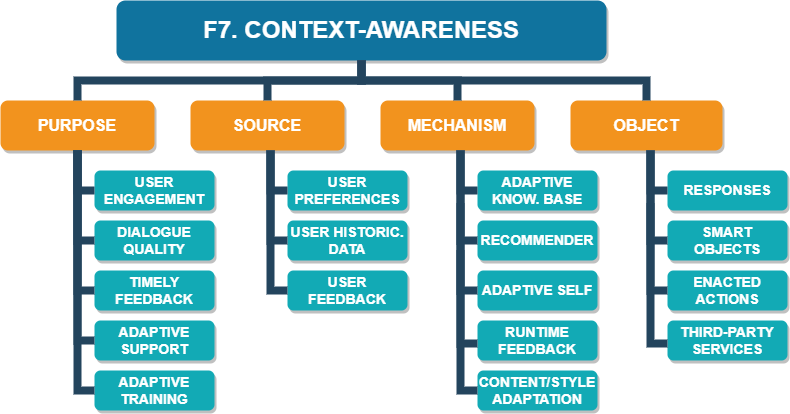}
  \caption{Aggregated results of conversational agents' context-awareness strategies in related literature}
  \label{fig:f7-sum}
\end{figure}

All in all, we identified 4 different dimensions in which context integration can be described: the \textbf{purpose} or main goal of the context integration strategy; the \textbf{source} or trigger of the adaptation; the \textbf{mechanism} used for integrating context into the conversational experience; and finally the \textbf{object} of adaptation, which is the entity or process affected and adapted by the context-aware knowledge. A categorization of the examples covered by the previous studies allows us to build and depict a general approach based on how context adaptation and personalization are integrated into conversational agents. Figure \ref{fig:f7-sum} is the result of such harmonization process.

\subsection{Findings of SQ3}

In this section, we summarize the main insights that can be extracted from the results of SQ3.

\noindentparagraph{\textbf{From a technical perspective, there are two major categories of conversational agents.}} We propose a classification of two major types or generations of conversational agents: \textit{deterministic} and \textit{\ed{ML-based}} approaches. We define each of these categories through a sub-set of 6 major characteristics based on the feature extraction process. These characteristics are summarized in Table \ref{tab:sq3}. 

\noindentparagraph{\textbf{Deterministic conversational agents are the first generation of dialogue systems.}} This type of conversational agents are designed as a fixed knowledge conversational base, which allows a deterministic mapping between user input and responses output from a closed data-set of responses. They are developed using standardized rule-based tools like pattern matching techniques \add{\cite{Adamopoulou2020-yj}}, which are reported as predominant in literature \add{\cite{Adamopoulou2020-yj, Abdul-Kader2015-jy,Hussain2019-lj}}. These strategies, while representing an early, first generation of conversational agents, provide an important advantage in terms of their simplicity, their maturity, their efficiency of development and the lack of need for training data.

\noindentparagraph{\textbf{\ed{ML-based} conversational agents are the second generation of dialogue systems.}} This type of conversational agents are designed as an adaptive knowledge conversational base, which is capable of processing and interpreting user input and either predict or create original generative output as a response to that user input \add{\cite{Safi2020-cy}}. These approaches are an emerging trend that serve from up-to-date machine learning and deep learning techniques (e.g., RNN, LSTM, Seq2Seq) which benefit from in-dialogue context integration \add{\cite{Bavaresco2020-md}} and self-adaptation \add{\cite{Janssen2020-fp}} to provide an adaptive conversational smart process. They require large data-sets and expert knowledge to be developed while offering a broader and more personalized communication experience with a higher specialization degree \add{\cite{Adamopoulou2020-yj}}. \addreview{Alternatively, the use of transformer deep learning models (e.g., GPT, DIET, BART), is becoming a pervarsive trend with potential benefits in terms of both efficiency and efficacy of advanced conversational experiences. These models have significantly contributed into building and delivering more efficient, scalable language models \cite{Brown2020}, depicting practical transformer-based architectures \cite{DIET2020} or even for additional, support tasks like efficient translation \cite{Hardalov2018}.}

\begin{table}[h]
\caption{General categorization of conversational agents design approaches based on F5 and F6}
\label{tab:sq3}
  \begin{tabular}{p{0.3\linewidth}p{0.3\linewidth}p{0.3\linewidth}}
    \toprule
    \textbf{Characteristic}&\textbf{Deterministic approaches}&\textbf{\ed{ML-based} approaches}\\
    \midrule
    \textbf{Behaviour}&Deterministic&Non-deterministic\\
    \textbf{Output}&Static&Dynamic\\
    \textbf{Knowledge base}&Fixed&Adaptive\\
    \textbf{Complexity}&Low&High\\
    \textbf{Maturity}&High&\ed{Medium}\\
    \textbf{Pervasiveness}&High&\ed{Medium/High}\\
    \bottomrule
  \end{tabular}
\end{table}

\noindentparagraph{\textbf{Deterministic and \ed{ML-based} are complementary approaches.}} The advantages and disadvantages of each approach can be used for choosing the best solution for a specific scenario, or even for the design and deployment of hybrid approaches. Hybrid design can serve from rule-based mechanisms for deterministic response generation in combination with generative capabilities and self-adaptive mechanisms through context-awareness to increase the degree of precision and accurateness of the conversational process \add{\cite{Adamopoulou2020-yj}}. Consequently, an accurate and more precise conversational process can be integrated while increasing the satisfaction of the user when interacting with the agent through additional characteristics like human or individual similarity (F4) and improving the performance in terms of quality evaluation in terms of task or functional effectiveness (F9).

\noindentparagraph{\textbf{Context-awareness is yet to be fully explored.}} There is a lack of synthesized, structured knowledge regarding context-awareness strategies for full-personalized conversational agents \cite{Bavaresco2020-md, De_Barcelos_Silva2020-gw}. The surveyed studies and the results of F7 can be used to identify 3 major research lines in terms of context-awareness strategies: (1) integrating additional data sources (e.g., user profiles \add{\cite{Knote2018-dx}}, third-party services \add{\cite{Kocaballi2019-my}}); (2) exploring and categorizing adaptive mechanisms (e.g., adaptive knowledge base \add{\cite{Janssen2020-fp}}, runtime feedback collection \add{\cite{Kocaballi2019-my}}), including HCI feature adaptation (e.g. communicational style, character/mood of the agent \add{\cite{Bavaresco2020-md}}); and (3) extend the scope of adaptation beyond the agent's response verbal content (e.g., connected smart objects \add{\cite{Knote2018-dx}}, third-party applications \add{\cite{Kocaballi2019-my}}).

%% file: content/3_4_training_testing_eval.tex
\section{Training, testing and evaluation (SQ4)}
\label{sec:sq4}

\subsection{Data-sets and data items (F8)}
\label{subsec:f8}

Despite being a critical issue for training, testing and evaluating conversational agents, information and discussion on data repositories, data-sets and data items is missing from almost all secondary studies included in this research. 
Given that dissertation is limited, and no formal categorization is provided, we propose a narrative approach to report and comment on this feature.

Dsouza et al. \cite{Dsouza2019-px} categorize data sources according to the domain-specific studies covered by their healthcare-focused research. Mainly, they identify three data sources: medical knowledge repositories, user information databases and conversation scripts. Medical knowledge repositories store domain-specific information data related to the area of application of the conversational agent. These include private or internal data repositories storing medical information like brain and facial images collected through traditional medical activity \cite{Vijayan2018-jk}, and public or online available sources like disease-specific documents (e.g., asthma) from Wikipedia or other public disease repositories \cite{Kadariya2019-vy}. User information databases are reported to be used especially for context adaptation and personalization tasks (as depicted in \ref{subsec:f7}), but no further insights are reported. Some examples include the use of demographic data \cite{Kadariya2019-vy} and patient’s electronic health record \cite{Miranda2019-mr}. Finally, a minority of studies use publicly available conversation scripts like call-centre and frequently asked question internal logs \cite{Abashev2017-rk,Wong2011-tp}. It is relevant to highlight that a significant amount of primary studies do not mention the type of data used for either training, testing or evaluation.

On the other hand, Safi et al. \cite{Safi2020-cy} identify and report the data-sets used by primary studies for evaluation analysis. Although most examples are healthcare-related, they also report commerce and generic-purpose conversational agents. In alignment with Dsouza et al. \cite{Dsouza2019-px}, healthcare-focused primary studies use public or private medical knowledge repositories like scrapped data from medical-related forums \cite{Belfin2019-sm} or disease-symptom mapping knowledge databases \cite{Rai2018-rs}. Two additional insights can be extracted from these examples. First, for generic design approaches, \addreview{there is a clear trend towards the use of} conversational scripts like Telegram \cite{telegram}, public chat repositories \cite{PerezSoler2018-gl} or the use of public, specific-purpose repositories like DBpedia \cite{dbpedia} or Stanford CoreNLP \cite{stanford} which provide academic data-sets for training, testing and evaluating in domain-specific contexts like commerce transactions \cite{Handoyo2018-as} or any closed-domain chatbot design \cite{Nigam2019-sp}. And second, as briefly mentioned by Adamopoulou and Moussiades \cite{Adamopoulou2020-yj}, the use of publicly available corpora reported by primary studies might be used for inferring training features and evaluation strategies in domain-related conversational agents.

\subsection{Quality and evaluation methods (F9)}

Alongside technical implementation specifications (F6), analysis on how research literature addresses the evaluation of conversational agents is one of the most frequent subjects of dedicated research, from a general overview of evaluation methods \cite{Kocaballi2019-my} to full-dedicated discussion \cite{Casas2020-ci} \add{and taxonomies of evaluation methods and metrics \cite{Ren_2019,Fitrianie2019-xs}}. We classify the discussion based on two different approaches: analysis on quality characteristics (i.e., \textit{what} is evaluated) and on evaluation methods and metrics (i.e., \textit{how} they are evaluated).

\begin{figure}[h]
  \centering
  \includegraphics[width=0.75\linewidth]{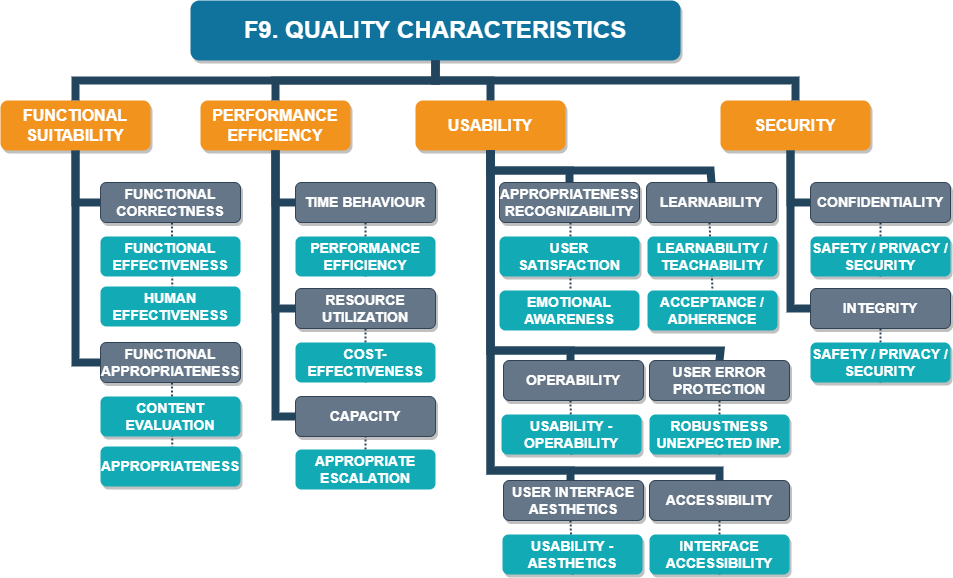}
  \caption{Aggregated results of conversational agents' quality characteristics in related literature}
  \label{fig:f9-1-sum}
\end{figure}

For quality characteristics, we propose a feature extraction process based on the ISO/IEC 25010 software product quality model \cite{iso25010}. We extract and compare the quality characteristics discussed and reported in the secondary studies covered by these features, and we align them with the ISO/IEC 25010 quality characteristics and sub-characteristics. This feature extraction process allows (1) to organize the evaluation of quality characteristics following a standardized, well-known framework, and (2) to exhaustively compare traditional software quality characteristics with the most frequent and relevant characteristics in the conversational agents' field, which allows us to detect and reflect on those quality characteristics which are reported as most important by the literature. Figure \ref{fig:f9-1-sum} summarizes the results of this alignment analysis. For simplicity, we only include those quality characteristics from ISO/IEC 25010 to which a match was found among the surveyed studies.

\begin{itemize}
    \item \textbf{Functional suitability}. The focus on functional suitability lies on the degree to which the interaction with the agent provides accurate responses with the required level of precision, as well as the degree to which these responses facilitate the achievement of users' goals and the tasks exposed by the agent. Regarding \textit{functional correctness}, the most common term in literature is \textit{effectiveness} \add{\cite{Milne-Ives2020-zl, Casas2020-ci, Radziwill2017-ws, Ren_2019, Guerino2020-vc}}. Casas et al. \cite{Casas2020-ci} differentiate between \textit{functional effectiveness}, which includes objective measures like command interpretation accuracy  and speech synthesis and generation performance \add{\cite{Coniam_2014}}, and \textit{human effectiveness}, which relates to the human similarity footing dimension as described in Section \ref{subsec:f4} \add{and for which the Turing test has been historically a standard reference for its evaluation \cite{Saygin_2000}, despite recent concerns regarding the relevance of the human effectiveness when interacting with chatbots \cite{Casas2020-ci}}. 
    Regarding \textit{functional appropriateness}, relevancy of content or \textit{content evaluation} is the most mentioned quality characteristic \add{\cite{Kocaballi2019-my, Maroengsit2019-by, Dsouza2019-px}}. As defined by Kocaballi et al. \cite{Kocaballi2019-my}, personalized content adaptation (e.g., topic suitability and verbal HCI features) is evaluated to achieve appropriateness in terms of users' goals achievement to match each user's specific needs \cite{Levin2006}. Milne-Ives et al. \cite{Milne-Ives2020-zl} specifically refer to \textit{appropriateness} of content and the generated responses, and it is reported as one of the most frequently used in the evaluation of conversational agents. 
    
    \item \textbf{Performance efficiency}. Few examples are mentioned which might be aligned to performance efficiency as defined in ISO/IEC 25010. The most common, shared term in this quality characteristic refers to the \textit{time behaviour} sub-characteristics, to which research generally refers to as \textit{performance efficiency} \cite{Casas2020-ci, Maroengsit2019-by}. In terms of \textit{resource utilization}, Milne-Ives et al. \cite{Milne-Ives2020-zl} \add{and Ren et al. \cite{Ren_2019}} report as a major quality characteristic \textit{cost-effectiveness} in the means of the relation between the cost (i.e., the resources) and the \textit{effectiveness} characteristic depicted before, \add{which can be evaluated through a compared experimental analysis between chatbot-assisted and traditional tasks in terms of task completion time \cite{Havik2019-wf}.}
    It is relevant to notice that Casas et al. \cite{Casas2020-ci} and Radziwill and Benton \cite{Radziwill2017-ws} use the concept of \textit{efficiency} in the context of usability in ISO standards as defined by Abran et al. \cite{Abran2003-ks}. Therefore, their insights and comments on this quality attribute must not be considered under this category. Regarding \textit{capacity}, Radziwill and Benton \cite{Radziwill2017-ws} report as an example of a quality attribute the \textit{appropriate escalation} of channel services for assisting user traffic requests.
    
    \item \textbf{Usability}. \ed{Some studies \cite{Radziwill2017-ws, Ren_2019}} propose to align the quality characteristics collected from primary studies with the concept of usability reported in ISO 9241 \cite{Abran2003-ks} in terms of efficiency \add{\cite{galitsky-ilvovsky-2017-chatbot}}, effectiveness and satisfaction \add{\cite{hancock-etal-2019-learning}}. Regarding the latter, \textit{user satisfaction} or simply \textit{satisfaction}, which relates to the degree to which users acknowledge the usefulness of the conversational agent to assist their needs (i.e., \textit{appropriateness recognizability}), is reported as a major quality characteristic by several secondary studies \cite{Kocaballi2019-my, Milne-Ives2020-zl, Casas2020-ci}. Additionally, other studies refine the concept of \textit{satisfaction}. As defined by Kocaballi et al. \cite{Kocaballi2019-my}, \textit{emotional awareness} of the agent is also identified as a quality attribute to be measured and which contributes to overall \textit{satisfaction}. Hobert \cite{Hobert2019-gl} reports further psychological factors related to this \textit{satisfaction} characteristic like user enjoyment, which is supplemented by Pérez et al. \cite{Perez2020-xt} with other HCI-related attributes like humanity, affection or friendliness of the agent.
    On a secondary basis, \textit{learnability} or \textit{teachability} of the conversational agent is also a literal concept from the ISO/IEC 25010 covered by research literature \add{\cite{Ren_2019, Guerino2020-vc, Perez2020-xt}}, including synonyms or related approaches like \textit{acceptability} as defined by Milne-Ives et al. \cite{Milne-Ives2020-zl}. Remaining sub-characteristics reported in Figure \ref{fig:f9-1-sum} are not generally discussed among secondary studies. Sub-characteristics like \textit{operability}, \textit{user interface aesthetics} and \textit{accessibility} are superficially discussed and covered by the concept of \textit{usability} as defined in some studies \cite{Hobert2019-gl}. Finally, it is relevant to highlight a special mention by Radziwill and Benton \cite{Radziwill2017-ws} concerning \textit{user error protection} sub-characteristic, which is reported as \textit{robustness of unexpected input} and identified as the capacity of the agent to avoid inappropriate utterances and apply damage control techniques when manipulation or unexpected input messages are sent \add{\cite{Kenny_2011}}.
    
    \item \textbf{Security}. Finally, security sub-characteristics like \textit{confidentiality} and \textit{integrity} are partially covered by the major category as reported by Milne-Ives et al. \cite{Milne-Ives2020-zl}, which is generally summarized into \textit{safety / privacy / security} concerns. These terms, which are sometimes referred to as synonyms during evaluation, relate to the more restricted concept of \textit{trustworthiness} (from a security point of view) reported by Pérez et al. \cite{Perez2020-xt}. The \textit{confidentiality} characteristic is emphasized on both subjective perceptions and objective evaluations on privacy concerns, while \textit{integrity} focus on data operation and management (e.g., user private data \add{\cite{Meinert_2018}}, manipulation and modification of such data \add{\cite{Thieltges_2016}}).
\end{itemize}

\begin{figure}[h]
  \centering
  \includegraphics[width=0.90\linewidth]{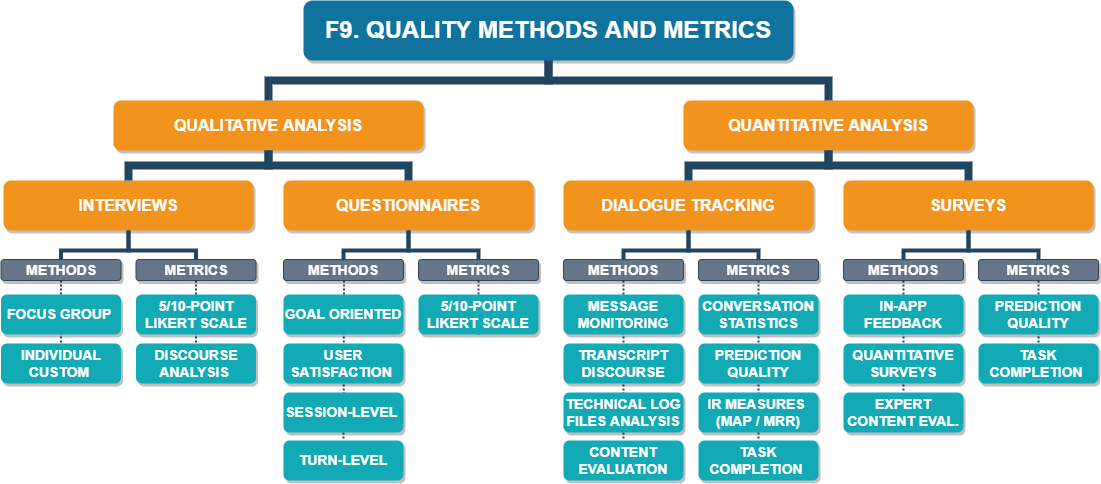}
  \caption{Aggregated results of conversational agents' evaluation methods and metrics in related literature}
  \label{fig:f9-2-sum}
\end{figure}

For the analysis of quality methods and metrics, we suggest a categorization between \textbf{qualitative} and \textbf{quantitative} analysis methodologies. 
Figure \ref{fig:f9-2-sum} summarizes different methododologies covered by secondary studies, as well as a list of frequently used metrics for each of these general categories.

\begin{itemize}
    \item \textbf{Qualitative analysis}. We identify two main methodologies for qualitative evaluation: \textit{interviews} and \textit{qualitative questionnaires}.
    \begin{itemize}
        \item \textit{Interviews}. As defined by Hobert \cite{Hobert2019-gl}, qualitative interviews allow evaluators to get detailed feedback from the participants of an experiment with the conversational agent in regard to its impact and the effects of user interaction with the agent. Based on Kocaballi et al.  \cite{Kocaballi2019-my}, we observe two different patterns for conducting interviews, based on the primary studies covered in their research: \textit{focus group interviews}, where participants might be grouped into small groups to facilitate discussion and the exchange and contrast of opinions and impressions \cite{Rhee2014-am}; and \textit{individual custom interviews}, which are carried on separately for each individual based on their experience with the agent and their personal user experience as in open-ended, semi-structured interviews \cite{Sillice2018-mf}. 
        The most frequent metric is the use of a \textit{Likert scale} measure for objective questions (e.g., measuring the level of satisfaction of a specific response \cite{Sillice2018-mf}), alongside \textit{discourse analysis} on interview transcripts \add{\cite{Braun_2017}, restrospective think-aloud sessions \cite{Chen_2018} or cognitive walkthrough \cite{Saenz_2017}. These techniques allow a metric-based evaluation of quality characteristics like \textit{learnability/teachability}, \textit{acceptance/adherence} and \textit{user satisfaction} \cite{Ren_2019}}.  
        
        \item \textit{Questionnaires}. Two different dimensions are extracted from research literature for evaluation through \textit{qualitative questionnaires}. The first one relates to purpose of the questionnaire, for which in the results reported by Kocaballi et al. \cite{Kocaballi2019-my} we can differentiate between \textit{goal oriented questionnaires} and \textit{user satisfaction questionnaires}. \textit{Goal oriented questionnaires} are designed to measure and monitor features or qualities to which the interaction with the agent has a significant impact. For instance, in the healthcare domain, questionnaires can be used for measuring disease symptoms like depression or anxiety before and after interaction with a conversational agent focused on dealing and treating these pathologies \cite{Fulmer2018-jg}. On the other hand, \textit{user satisfaction questionnaires} cover the topics related to the \textit{usability} quality characteristics reported in Figure \ref{fig:f9-2-sum}, including examples like emotional awareness, learning and relevancy of content \cite{Fitzpatrick2017-dv}. 
        Additionally, Maroengsit et al. \cite{Maroengsit2019-by} differentiate two levels for user satisfaction evaluation: \textit{session-level questionnaires}, where users are asked to evaluate an entire conversation session with the agent based on a different set of factors (e.g., appropriateness, empathy, helpfulness \cite{Xu2017-lg}); and \textit{turn-level questionnaires}, where users must evaluate each response from a conversational agent, which might be used for an average rating based on user or expert evaluation \cite{Kazi2012-cm}. \add{Beyond ad-hoc questionnaire design, some examples of commonly used questionnaires for qualitative analysis include the \textit{System Usability Scale} (SUS), which is typically combined with other evaluation techniques like direct observation \cite{Preece_2017}, and the \textit{NASA Task Load Index (NASA-TLX)}, which allows a subjective evaluation of the perceived reduced workload when performing agent-assisted tasks \cite{Guerino2020-vc}.}
        As qualitative interviews, the use of a 5/10-point \textit{Likert scale} is the most frequent metric used in questionnaires.
    \end{itemize}
    
    \item \textbf{Quantitative analysis}. We identify two main methodologies for quantitative evaluation: \textit{dialogue tracking} and \textit{quantitative surveys}.

    \begin{itemize}
        \item \textit{Dialogue tracking}. Automatic or semi-automatic monitoring and analysis of the user-agent communicational process (i.e., the natural language messages sent by the user and the responses generated by the chatbot) is considered a key evaluation method for both \textit{usability} and \textit{functional suitability} quality evaluations \cite{Kocaballi2019-my, Masche2018-sw, Hobert2019-gl}. 
        \textit{Message monitoring} techniques for meta-data features regarding the user-agent interaction is one of the most common methods, given its simplicity and the relevance of the data in terms of analysing how users interact with agents, which guides the evaluation of quality attributes like \textit{user engagement} \cite{Fulmer2018-jg}. In addition, more advanced methods like \textit{transcript discourse analysis} \add{\cite{Hobert2019-gl}}, \textit{technical log files analysis} \add{\cite{Radziwill2017-ws}} and \textit{content evaluation} \add{\cite{Maroengsit2019-by}} are used to get deeper insights both into the technical correctness of the agent and its performance. These automatic methods are typically measured through standard numerical forms including \textit{prediction quality measures} (e.g., accuracy, precision, recall and F-measure) or \textit{information retrieval measures} (e.g., mean average precision and mean reciprocal rank) \add{for \textit{functional suitability} evaluation \cite{Maroengsit2019-by, Ren_2019}}. These metrics are especially relevant for functional evaluation in context-specific or Q\&A conversational agents, for which expert manual annotation process of the agent's responses can be used for the evaluation measure \cite{Cui2017-pn}. Additionally, \textit{conversation statistics}  (e.g., number of messages sent by user, length of these messages and number of chatbot responses rated as positive) and \textit{task completion measures} \add{(e.g., task completion rate, task completion time) are typically used for \textit{performance efficiency} evaluation \cite{Perez2020-xt, Radziwill2017-ws, Ren_2019}}.
        
        \item \textit{Surveys}. With the focus on \textit{functional suitability}, quantitative surveys allow involving users in the annotation process for evaluating the quality of the responses generated by the agent and the level of achievement of user's tasks and goals. As reported by Kocaballi et al. \cite{Kocaballi2019-my}, the use of \textit{in-app feedback questions} is a relevant example to progressively measure the task completion ratio or the number of messages evaluated as errors or bad quality answers \cite{Harper2008-nm}. The use of \textit{pre or post quantitative surveys} can be used to the same end \add{\cite{Hobert2019-gl, Ren_2019}}, with the ability to provide quantitative layouts before and after the session, which can later be used to compute and measure the level of task or goal achievement, e.g., as in the learning success of an educational chatbot for a specific skill or state of knowledge \cite{Hobert2019-gl}. Finally, Maroengsit et al. \cite{Maroengsit2019-by} and Fitrianie et al. \cite{Fitrianie2019-xs} highlight the use of \textit{expert content evaluation} surveys to facilitate the annotation process for expert evaluation of the response or even the ranked set of response candidates of a conversational agent, which is used to report \textit{prediction quality metrics} like top-N response accuracy or task completion measures \cite{Qiu2017-lw}. 
    \end{itemize}
\end{itemize}

\add{To support the design and execution of evaluation plans for conversational agents, industry and academia are recently devoted to building methodologies and tool-supported solutions for semi-automatic, assisted integration of some of the aforementioned evaluation methods and metrics. While earlier approaches introduced some groundwork for specific quality characteristics (e.g., a framework for chatbot security evaluation \cite{Bozic_2018}), more recent contributions to the field introduce methodologies and frameworks for a holistic quality evaluation, like BoTest \cite{Ruane_2018}, Bottester \cite{Vasconcelos_2017}, and a framework for metamorphic testing \cite{Bozic_2019}. With respect to software-based solutions, a reference in the academic field is BOTIUM \footnote{https://www.botium.ai/}, a suite of open-source components for automated chatbot evaluation. BOTIUM allows integration with some of the major software-based solutions for chatbot development (e.g., Google DialogFlow, Microsoft LUIS) and has served as background infrastructure for more advanced evaluation methodologies and frameworks like CHARM \cite{BravoSantos_2020}. Other related alternatives include Haptik \footnote{https://www.haptik.ai/}, BotAnalytics \footnote{https://botanalytics.co/} and ChatbotTest\footnote{https://chatbottest.com/}. These tools allow designing specific automated tests, including full end-to-end user flow simulation, automated answer accuracy measures, stress-testing and user interaction monitoring.}

\subsection{Findings of SQ4}

In this section, we summarize the main insights that can be extracted from the results of SQ4.

\noindentparagraph{\textbf{There is a research gap on data-set and data-items dissertation.}} While building knowledge bases for deterministic conversational agents has been traditionally addressed through the dedicated work of domain experts (e.g., elaboration of pattern matching templates \add{\cite{Adamopoulou2020-yj, Abdul-Kader2015-jy}}), state-of-the-art solutions involving \ed{ML-based} techniques require a large amount of data
to train the models used in these solutions \add{\cite{Nuruzzaman2018-iv}}. Consequently, dimensions such as type, format, quality and amount of data items must be researched from a general perspective to address the challenge of gathering data for the training of conversational agents' knowledge bases. To this end, we suggest to explore and follow the examples of primary studies using public, open-source data repositories of NLP data-sets for dialogue tasks, \add{as suggested by related work in the field \cite{Adamopoulou2020-yj}}. Research should primarily focus on data-driven approaches in the context of novel technical solutions like the use of recurrent neural networks and context-aware conversational solutions, which face specific challenges in terms of type and amount of available data. 

\noindentparagraph{\textbf{Quantitative evaluation is mainly covered by dialogue tracking methods.}} These techniques are limited to evaluation methodologies for low-level sub-tasks of the conversational process, including intent and entity classification \add{\cite{Maroengsit2019-by}} and natural language response generation correctness \add{\cite{Perez2020-xt}}. On the other hand, qualitative analysis evaluation is predominantly oriented towards quality characteristics focused on the capacity of conversational agents to accurately match and suit the specific goals and needs of their users \add{\cite{Kocaballi2019-my}}. Furthermore, quality sub-characteristics like appropriateness recognizability (e.g., \textit{user satisfaction}) or learnability (e.g., \textit{acceptance/adherence}) involving the perception of the user are widely addressed and incorporated into most quality evaluation strategies \add{\cite{Kocaballi2019-my, Milne-Ives2020-zl, Hobert2019-gl, Ren_2019}}. 

\noindentparagraph{\textbf{Qualitative evaluation is designed as a user-oriented method based on subjective metrics.}} Concepts like \textit{satisfaction} or \textit{awareness} are repeatedly discussed through literature to address different strategies for capturing the degree to which users perceive conversational agents as helpful tools, whether we refer to usability \add{\cite{Hobert2019-gl, Perez2020-xt}}, performance suitability \add{\cite{Milne-Ives2020-zl, Guerino2020-vc}} or even reliability \add{\cite{Zierau_2020}}. However, secondary studies do not provide further details on the design criteria to design and carry on these types of experiments. Given the relevance and the impact that HCI features have demonstrated to have on the user experience with the conversational agent, it remains obscure how these HCI features can be captured and evaluated into an interview or a questionnaire-based evaluation in order to process, model, interpret and compare conversational agents and human behaviours.

    
    
    

%% file: content/3_5_future_challenges.tex
\section{Research challenges in literature (SQ5)}
\label{sec:sq5}

\subsection{Research challenges (F10)}
\label{subsec:f10}

\add{In this section, we build the dissertation on the current research challenges collected through the findings of SQ1-SQ4. We use the surveyed studies as well as the inspected primary studies to extract and identify specific limitations in each SQ. These individual, research-focused gaps are compared and aggregated to build a holistic perspective on the main global research challenges reported by literature across all the SQ covered in this survey.} 

\add{Based on this analysis,} we synthesize and report 6 major challenges in the field of conversational agents, which we envisage that will guide future research:

\begin{enumerate}
    \item \textbf{\ed{Increase the pervasiveness of conversational agents in concrete domain-specific and target-specific areas.}} Advanced research in domain-specific scenarios has deeply explored the implications of the design specifications in terms of HCI features and technical development. De Barcelos et al. \cite{De_Barcelos_Silva2020-gw} unveil a recent trend in integrating conversational agents into healthcare (e.g., vital signs monitoring, mental health), education or daily-life scenarios. However, they highlight that the articles show that most people only use conversational agents for simple tasks. A similar conclusion is reported by Syvänen and Valentini \cite{Syvanen2020-nb}, as they report as a major future work among primary studies to improve the proof-of-concept conversational agents to become real tools for their respective domains. Similar considerations are made regarding specific target audiences (e.g., elderly, functional diversity) \cite{Kocaballi2019-my, VanPixteren2020-zm}. 
    Fully exploiting the potential of some specific scenarios and the characteristics of specific target audiences in real environments remains to be thoroughly explored.
    
    \item \textbf{Improve user engagement through perceived quality.} Whereas user communication and user experience is the core of the conversational process, analysing and adapting HCI features to improve user engagement is still a major key challenge. In fact, Syvänen and Valentini \cite{Syvanen2020-nb} claim that the concept of \textit{engagement} is understudied, and therefore the consequences and the impact on user engagement concerning the agent's communicational capabilities must be researched with further details. De Barcelos et al. \cite{De_Barcelos_Silva2020-gw} focus on user engagement through the need for improving the perceived quality in terms of usability of the agent and user satisfaction, but also in terms of the perceived security and privacy of the data shared with the agent. Highly advanced taxonomies provide a theoretical background for the categorization of relevant HCI features and their impact on these features. Based on the knowledge extracted from these taxonomies, users' experience must be investigated from the user's perception to improve the perceived quality through the conversational process.
    
    \item \textbf{\ed{Extend empirical research on the impact of verbal and non-verbal HCI features towards individual similarity.}} Concerning HCI features, the aggregated knowledge depicted in Figure \ref{fig:f4-sum} allows us to identify relevant research gaps in terms of HCI features according to the modality and the footing of such features. Van Pinxteren et al. \cite{VanPixteren2020-zm} report the need for improving communicative behaviours, with special emphasis on non-verbal (e.g., speech rate) and physical (e.g., gestures) features. In alignment with the integration of context-aware and personalization techniques (F7), individual similarity \add{(as defined in Section \ref{subsec:f4})} can be further investigated to improve user engagement through the adaptation of the communicational process. To be specific, response content customization and the adaptation of non-verbal features like auditory or appearance features are some areas where we suggest that research should focus in the near future.
    
    \item \textbf{\ed{Design and develop mechanisms for the construction of fully-personalized agents through user profiling and context-awareness techniques.}} \ed{ML-based} technologies like RNNs\addreview{, transformer models} and recent examples of context-aware conversational agents through the extension of the conversation scope are the fundamental base for future research to achieve full personalization \addreview{not only in terms of functional appropriateness, but also in terms of performance efficiency}. Bavaresco et al. \cite{Bavaresco2020-md} report a few examples of how chatbots address self-learning, personalization and context-awareness in domain-specific scenarios, which can be generalized to high-level techniques like user profiling or user \ed{goals} analysis. De Barcelos et al. \cite{De_Barcelos_Silva2020-gw} also report personalization, user profiling, context understanding and user goals automatic recognition. \add{Beyond general strategies and methods for assistant personalization, the need for high-level mechanisms and tools for their construction is also a current challenge, for which software-based solutions of reference in the field (e.g., RASA) already outline that will guide the development of future versions in the next years \cite{rasa_2021}.} We envisage that future research will require the design and definition of clear, structured general practices \add{and tools} integrating user profiling techniques for autonomous user \ed{goals} detection, the adaptation of the communicational process (as suggested in SQ2) and the integration of contextual data \add{for such purpose}.
    
    \item \textbf{\ed{Synthesize and conduct structured research in the area of data sources, data types and data management for training and testing conversational agents.}} Literature reports a significant gap in terms of shared, structured knowledge in regard to data sources and data items for the training and evaluation of conversational agents. Bavaresco et al. \cite{Bavaresco2020-md} explicitly report that some domains have very limited knowledge base data sources for building and training the models in \ed{ML-based} solutions. A clear taxonomy of available data repositories based on a set of classification criteria (e.g., domain, purpose, size, type, format, reported evaluation quality, average meta-data features like messages length or speech rate) should help researchers focusing on these tasks as a starting point to evaluate and discuss the expectations of data availability and data suitability for their own research.
    
    \item \textbf{Extend qualitative analysis methodologies to improve perceived quality and appropriateness quality attributes.} User's perceived quality is an essential characteristic in conversational agents' quality evaluation. These quality characteristics focusing on perception and appropriateness (i.e., functional appropriateness, appropriateness recognizability) are \ed{deeply discussed in the surveyed studies}. Milne-Ives et al. \cite{Milne-Ives2020-zl} claim the need for extending qualitative evaluations to clarify and evaluate user quality perception, and consequently user engagement. Additionally, they highlight the need for identifying all the structural, physical and psychological barriers in these evaluation methodologies to identify and recognize how to improve user engagement and the penetration of conversational agents. 
    Consequently, we suggest that further research in the field of evaluation should be primarily focused on standard practices for qualitative analysis towards perceived quality. 
    
    \item \add{\textbf{Develop automated methodologies and tools for quantitative evaluations with advanced testing features.} In addition to the scarce availability of support tools for evaluating chatbots \cite{BravoSantos_2020}, there is a need for extending these tools with advanced testing features for the next generation of conversational agents, which include context-aware and personalization mechanisms \cite{Bavaresco2020-md, Zierau_2020}. In addition, other complex testing features like extended automatic textual data-set generation \cite{Jin_2020} and support for continuous integration \cite{BravoSantos_2020} will also be necessary to integrate into these methods and tools.}
    
\end{enumerate}

\subsection{Findings of SQ5}

In this section, we summarize the main insights that can be extracted from the results of SQ5.

\noindentparagraph{\textbf{User perceived quality is a key research challenge.}} Research challenges (1), (2), (3) and (6) are related to the degree to which users perceive conversational agents as useful tools. The future evolution of the pervasiveness of these systems is mainly conditioned by the insights and contributions of future research in terms of improving user experience through an advanced, efficient and accurate smart conversational process.

\noindentparagraph{\textbf{Context-awareness, personalization and data-driven research requires from structured, synthesized knowledge.}} Research challenges (4), (5) \add{and (7)} are related to the degree to which conversational agents are capable of building customized and personalized knowledge bases to provide fully-personalized experiences. As previously reported, there is a lack of structured, synthesized knowledge in these areas, \add{as well as complex, advanced tools and mechanisms for their design and construction}. Consequently, identifying and exploiting data sources and data items is a key challenge not only for the design of these systems, but also for its evolution and personalization.

    

%% file: content/5_limitations.tex
\section{Limitations}
\label{sec:limitations}

We evaluate the limitations based on the threats to validity as defined by Wohlin et al. \cite{Wohlin2012}.

\textbf{Construct and internal validity.} There are potential threats driven by the defined research strategy and the study selection process that might have led to miss or exclude relevant studies for the review\addreview{, or even to miss relevant venues in the dialogue systems research field}. Concerning missing secondary studies \addreview{or venues}, the data sources and additional reference strategies have been carefully designed and extended according to high quality assessment standards in terms of studies' coverage \cite{DARE5} and in adaptation to the software engineering field to mitigate as much as possible the risk of missing relevant studies. The search strings used for \textit{conversational agent} synonyms have been updated through a preliminary literature analysis and validated by the research method findings and the set of synonyms used by the secondary studies. Concerning the threat of excluding relevant studies, inclusion criteria have been designed using broad, flexible definitions of the terms \textit{conversational agent} and \textit{literature review}. On the other hand, domain-specific exclusion criteria (i.e., EC1, EC2) have been designed as strong restrictions to reduce the probability of excluding studies whose conclusions could be applied at any level to our research field. Finally, the risk of personal bias during the inclusion/exclusion criteria evaluation, the quality assessment or even during the feature extraction is also a possible threat. The first two processes have been reviewed and discussed among the authors to reduce personal bias by reaching a common agreement through discussion on the threats and limitations of the decisions taken at each evaluation. Concerning the feature extraction process, dedicated collaborative sessions with all the co-authors were carried out for each feature in alignment with each research questions to discuss the quality, reliability and potential threats of the results.

\textbf{External validity.} Missing relevant studies also apply to external validity for those papers whose full-text was not available even under explicit request to the authors (EC5), for which we did not receive any response. For those studies that were finally included in the study selection data-set, there is also a potential threat in terms of incomplete or inaccurate research information and conclusions reported by those secondary studies, especially for those with a low quality assessment score. Given that a conservative strategy was adopted to mitigate the construct validity threat of missing relevant studies, it was necessary to carefully inspect and trace all secondary study results which have been reported and integrated into our feature extraction and discussion results. Consequently, all features, examples, categories, methodologies and techniques have been linked to specific references to mitigate the risk of replicating inaccuracy on the results.

\textbf{Conclusion validity.} Personal bias in inclusion/exclusion evaluation, quality assessment and feature extraction processes might also lead to a validity threat in terms of the discussion and conclusions reported in this research, for which the same mitigation strategies can be applied. In terms of replicability of the study and its conclusions, the main threat is the restricted time span of the study search, which we have reduced by removing all time restrictions from the research strategy except for the date in which the search was executed. Despite this, assuming the inevitability of the time dimension consequences, the replication package attached to this study aims to provide an exhaustive, detailed summary of the entire research method execution to provide a clear, transparent overview of the research process and how it was conducted.

%% file: content/6_conclusions.tex
\section{Conclusions}
\label{sec:conclusions}

The integration of conversational agents in traditional software-based systems as intuitive, easy-to-use interfaces between humans and machines is expected to be a breakthrough in the future of HCI. As an emerging research field, academic research in the last years has focused on evaluating the specifications and the impact of conversational agents in a wide variety of contexts of use and domains. Research showcases the interdisciplinarity of conversational agents not only in terms of domains, but also in terms of scientific areas contributing to the state of the art of the field. Scientific and social areas like healthcare and education have significantly contributed to extend the applicable knowledge to the HCI field. And while laying the basis for the successful design of a smart conversational process in terms of user satisfaction, they also contribute to guide future research for the challenges that will require full attention in terms of user engagement, individual personalization or context-awareness strategies. Complementarily, commercial solutions are already exploring and exploiting the benefits of these software-based dialogue systems through successful solutions like smart-home assistants, autonomous call-centres or customer-service chatbots.

To this end, the latest contributions in the field benefit from the potential of deep learning solutions. These techniques allow effective integration of immediate, conversation-related knowledge using past interactions with the user through the use of recurrent neural networks. But adapting and personalizing the communicational process based on past interactions is just the beginning of full context-aware, personalized conversational agents. A holistic design of the data sources and adaptation mechanisms in a software system that can be integrated into a conversational agent is essential for designing customized solutions exploiting as much as possible the available knowledge for a better user experience. Consequently, effective design of data-driven solutions for the exploitation of contextual data in the conversational process is a key factor for future research.


All in all, this survey aims to provide a holistic, clear overview of the state of the art of the conversational agents' research field through a set of distributed taxonomies for the research disciplines covered by each SQ. These contributions are intended not only to validate and summarize the reported results, but also to build a clear exposition which might help to synthesize existing knowledge in the field and to those for which conversational agents are a scientific topic outside their research areas. Finally, we expect that these contributions also help to guide future research towards a new era of HCI through the successful integration of conversational agents.

%% file: content/A_acknowledgments.tex
\begin{acks}
With the support from the Secretariat for Universities and Research of the Ministry of Business and Knowledge of the Government of Catalonia and the European Social Fund.
The corresponding author gratefully acknowledges the Universitat Politècnica  de Catalunya and Banco Santander for the financial support of his predoctoral grant FPI-UPC. 
This paper has been funded by the Spanish Ministerio de Ciencia e Innovación under project / funding scheme PID2020-117191RB-I00 / AEI/10.13039/501100011033.

\begin{figure}[h]
  \centering
  \minipage{0.32\textwidth}
      \includegraphics[width=\linewidth]{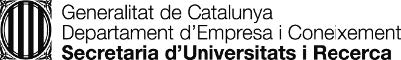}
  \endminipage
  \hfil
  \minipage{0.32\textwidth}
      \includegraphics[width=\linewidth]{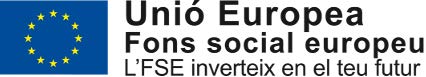}
  \endminipage
\end{figure}

\end{acks}